\pdfoutput=1

\documentclass[11pt]{article}

\usepackage[preprint]{acl}

\usepackage{times}
\usepackage{latexsym}
\usepackage{comment}
\usepackage{booktabs}
\usepackage{multirow}
\usepackage{longtable}
\usepackage{float}
\usepackage{placeins}
\usepackage[T1]{fontenc}

\usepackage[utf8]{inputenc}

\usepackage{microtype}

\usepackage{inconsolata}

\usepackage{graphicx}
\usepackage{caption} 
\usepackage{subcaption}
\usepackage{amsfonts}
\usepackage{amsmath}

\usepackage{algorithm}
\usepackage{algpseudocode}
\floatname{algorithm}{Hypothesis}

\usepackage{xcolor}
\definecolor{goldenGroupColor}{RGB}{235, 173, 162} 
\definecolor{controlGroupColor}{RGB}{73, 145, 157} 

\makeatletter
\renewcommand{\ALG@name}{Hypothesis}

\newenvironment{breakablealgorithm}
  {
   \begin{center}
     \refstepcounter{algorithm}
     \hrule height.8pt depth0pt \kern3pt
     \renewcommand{\caption}[2][\relax]{
       {\raggedright\textbf{\ALG@name~\thealgorithm} ##2\par}%
       \ifx\relax##1\relax 
         \addcontentsline{loa}{algorithm}{\protect\numberline{\thealgorithm}##2}%
       \else 
         \addcontentsline{loa}{algorithm}{\protect\numberline{\thealgorithm}##1}%
       \fi
       \kern3pt\hrule\kern3pt
     }
   \kern2pt 
  }{
     \kern2pt\hrule\relax
     \end{center}
     \kern2pt 
  }
\makeatother

\title{A Peek into Token Bias: Large Language Models Are Not Yet 
\\ Genuine Reasoners}

\author{Bowen Jiang\textsuperscript{1, 2}, Yangxinyu Xie\textsuperscript{1, 2}, Zhuoqun Hao\textsuperscript{1}, Xiaomeng Wang\textsuperscript{1}, \\ {\bf Tanwi Mallick\textsuperscript{2}, Weijie J. Su\textsuperscript{1}, Camillo J. Taylor\textsuperscript{1}, Dan Roth\textsuperscript{1}} \\  
\quad\ \ University of Pennsylvania\textsuperscript{1} \quad\ Argonne National Laboratory\textsuperscript{2} \\
Philadelphia, PA, 19104, USA \quad\quad Lemont, IL, 60439, USA \\
        {\tt\small \{bwjiang@seas, xinyux@wharton, zhuoqunh@sas, xwang1@wharton\}.upenn.edu}, \\ \tt\small tmallick@anl.gov, \tt\small \{suw@wharton, cjtaylor@seas, danroth@seas\}.upenn.edu
        }

\begin{document}
\maketitle
\begin{abstract}
This study introduces a hypothesis-testing framework to assess whether large language models (LLMs) possess genuine reasoning abilities or primarily depend on token bias. We go beyond evaluating LLMs on accuracy; rather, we aim to investigate their token bias in solving logical reasoning tasks. Specifically, we develop carefully controlled synthetic datasets, featuring conjunction fallacy and syllogistic problems. Our framework outlines a list of hypotheses where token biases are readily identifiable, with all null hypotheses assuming genuine reasoning capabilities of LLMs. The findings in this study suggest, with statistical guarantee, that most LLMs still struggle with logical reasoning. While they may perform well on classic problems, their success largely depends on recognizing superficial patterns with strong token bias, thereby raising concerns about their actual reasoning and generalization abilities. 
Codes and data are open-sourced at \href{https://github.com/bowen-upenn/llm_token_bias}{https://github.com/bowen-upenn/llm\_token\_bias}.
\end{abstract}    
\section{Introduction}\label{sec: intro}


Large language models (LLMs) have achieved remarkable progress in understanding and generating human-like text, triggering growing interest in the LLMs' theory of minds~\cite{kosinski2023evaluating, jamali2023unveiling, bubeck2023sparks} and decision-making abilities~\cite{lyu2023faithful, prasad2023adapt, jiang2024multi, jiang2024multiagentvqaexploringmultiagent, xie2024wildfiregpt}. However, there is ongoing debate about whether LLMs possess genuine reasoning capabilities, as evidence suggests that the performance of LLMs on reasoning tasks is correlated with how much the input's semantic content
supports a correct logical inference \cite{dasgupta2022language, li2023deceiving}. Should valid reasoning be applied, such a correlation would not exist, since a genuine reasoner should be able to derive the correct inference regardless of the context.


In this paper, we formalize this observation and say that an LLM is subject to \textbf{token bias} in a reasoning task if systematic changes to some or all tokens in the task descriptions - while keeping the underlying logic intact - allow us to predict the direction of the shift in the model's output. \textbf{A strong token bias suggests that the model is relying on superficial patterns in the input rather than truly understanding the underlying reasoning task.} This could lead to brittle performance that fails to generalize well to novel examples and phrasings encountered in the wild, which could differ from the spurious patterns the model may have overfitted to during training.

\begin{figure}[t]
  \centering
    \includegraphics[width=\linewidth]{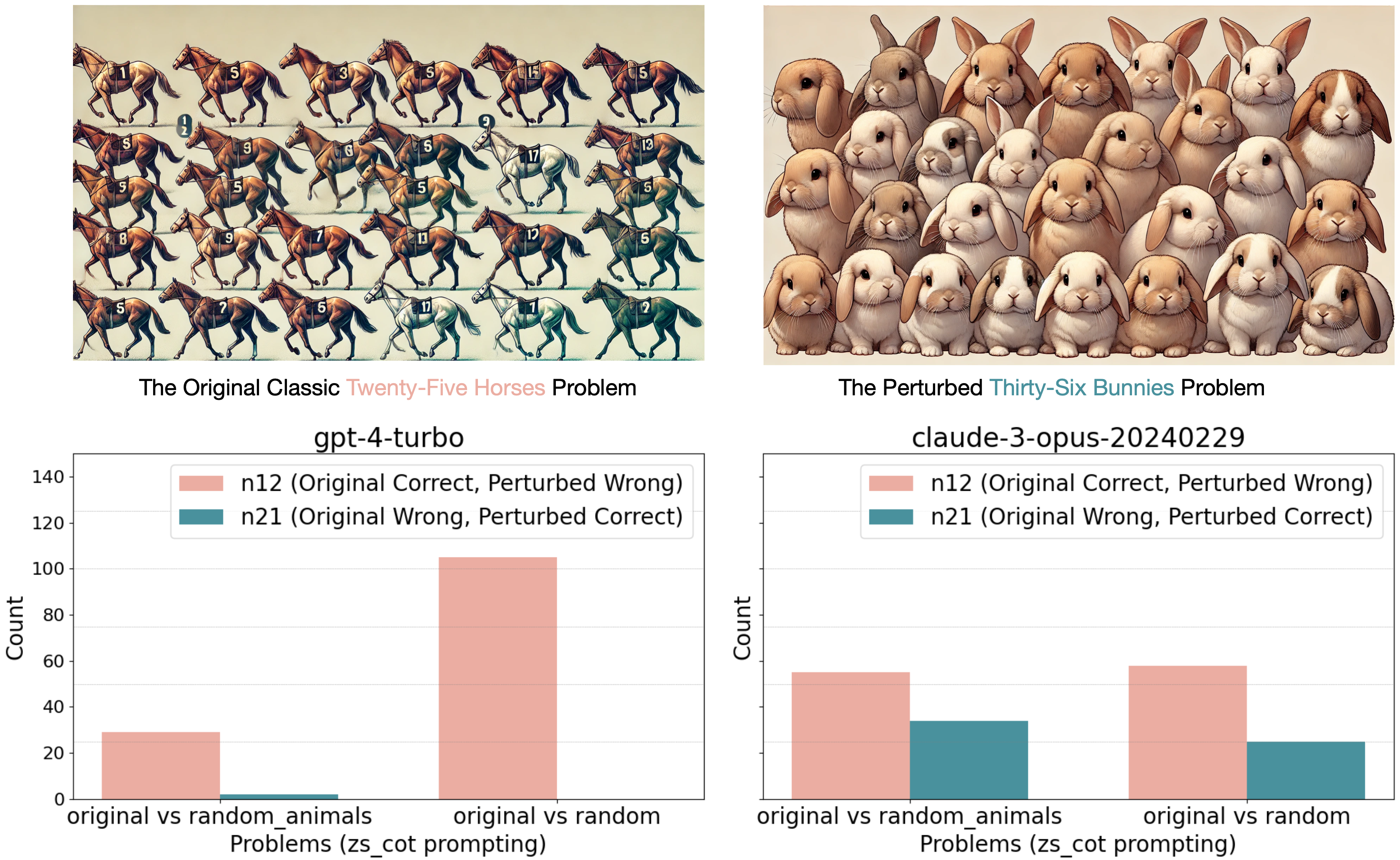}
    \caption{We illustrate token bias using the classic "twenty-five horses" problem in graph theory. The top two sub-figures, generated by GPT-4o for illustration purposes only\protect\footnotemark, demonstrate the concept by altering the name "horses" to "bunnies", irrelevant to the problem's underlying logic. The bottom two sub-figures show experimental results in GPT-4 and Claude, where performance significantly drops due to perturbations in animal names and numbers. In these plots, "Original" refers to the unaltered "twenty-five horses" problem, "random\_animals" alters only the animal names, and "random" alters both names and numbers. We observe $\textcolor{goldenGroupColor}{n12} > \textcolor{controlGroupColor}{n21}$ with statistical significance, meaning that there are more instances where \textcolor{goldenGroupColor}{the original problem is solved correctly while the perturbed problem is solved incorrectly}, compared to \textcolor{controlGroupColor}{the reverse}. As a result, our hypothesis testing confirms token bias in this scenario.}
    \label{fig:horses}
    \vspace{-3mm}
\end{figure}
\footnotetext{Interestingly, when we prompted GPT-4o to generate an image of "lop-eared bunnies", the model exhibited a visual token bias by depicting bunnies with four ears — both lop and erect — implying it associated the term "bunnies" with the presence of erect ears in images, without a genuine and logical understanding of a bunny's physical reality. We move the exploration of visual token bias to the future work.}


\begin{figure}[t]
  \centering
    \includegraphics[width=\linewidth]{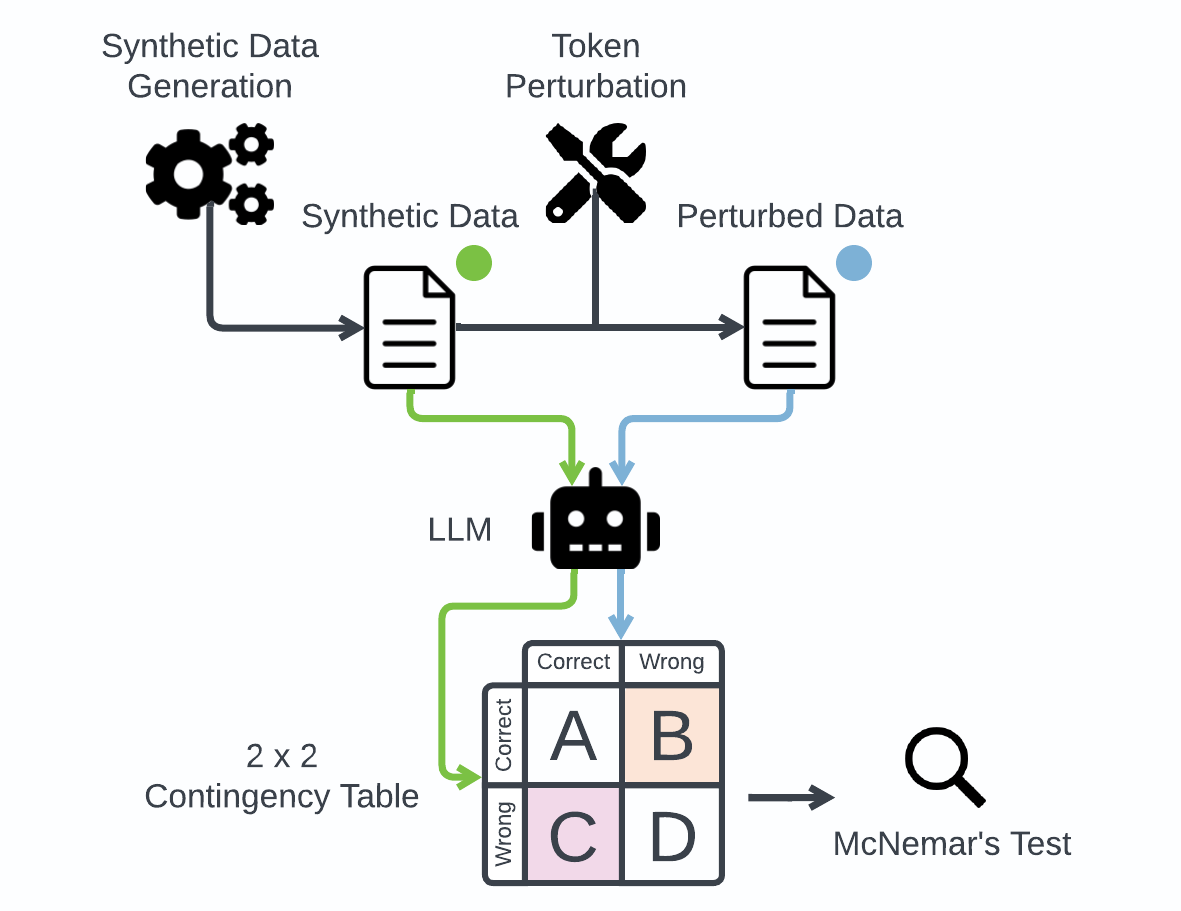}
    \caption{An illustration of the overall framework. We generate synthetic data, perform systematic token perturbations, and evaluate an LLM for comparative studies. The resulting contingency table, where A-D are integer values of counts, allows for subsequent statistical tests.}
    \label{fig:framework}
    \vspace{-3mm}
\end{figure}

\begin{figure}[t]
  \centering
    \includegraphics[width=\linewidth]{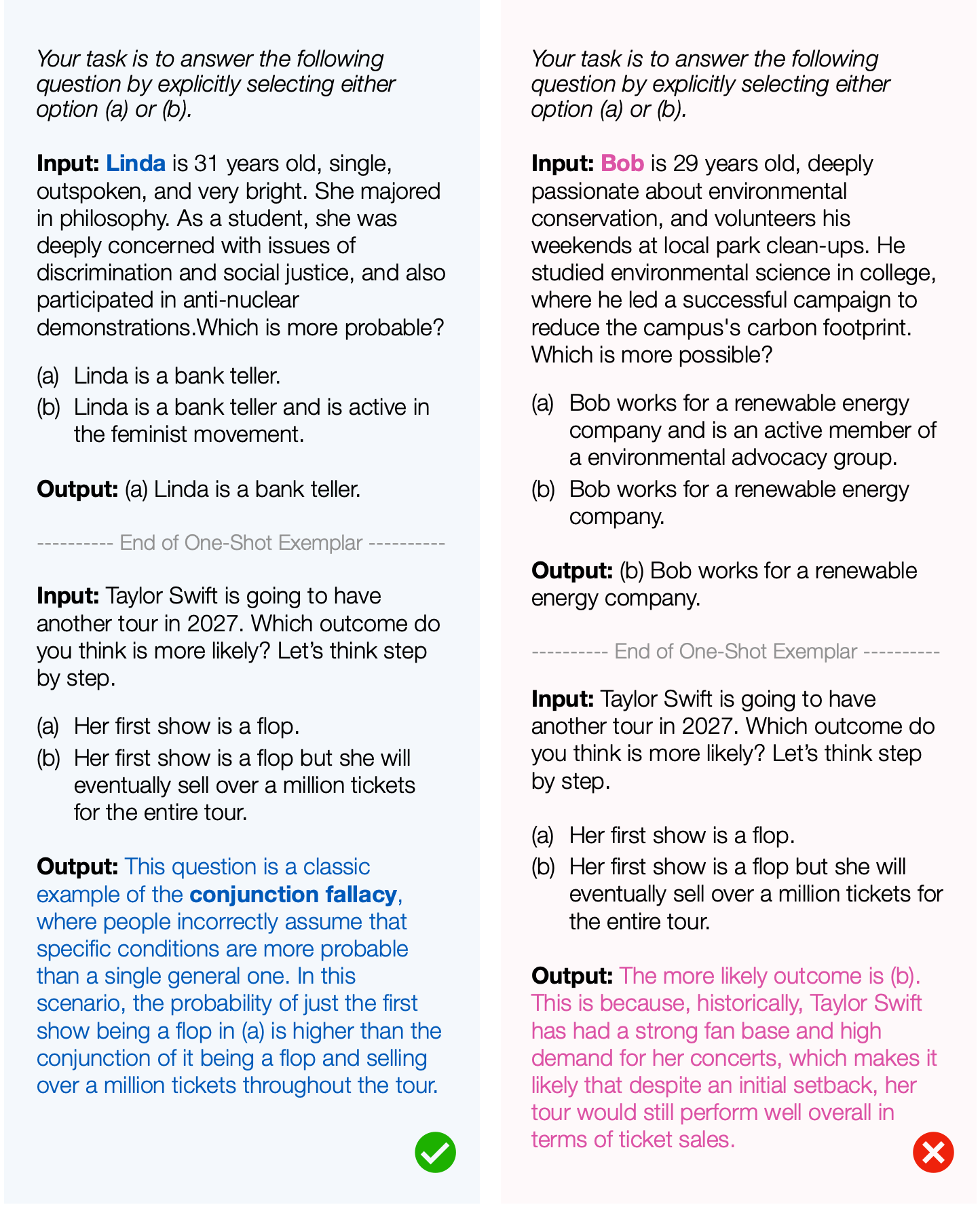}
    \caption{What is token bias? Here is another example exhibited by GPT-4. On the left, GPT-4 correctly identifies the conjunction fallacy and answers the question correctly, given the classical Linda Problem as the one-shot exemplar. On the right, however, the exemplar is rephrased by altering "Linda" to "Bob" while keeping the same logic, which surprisingly confuses the model.}
    \label{fig:bob}
    \vspace{-3mm}
\end{figure}


We explore several well-known logical fallacy problems from the cognitive science literature~\cite{tversky1983extensional, kahneman2011thinking}, which provide a clear playground for assessing the reasoning capabilities of LLMs. Figure~\ref{fig:horses} and \ref{fig:bob} depict two kinds of token biases found in our testing framework, where the model may be overfitting to specific tokens commonly found in classic problem statements. Since we observe many cases where state-of-the-art LLMs like GPT-4~\cite{achiam2023gpt} successfully identify logical fallacies under certain settings, we highlight the urgent need for a framework to tease out whether LLMs apply genuine reasoning or merely exploit token bias for their improved performance.

\textbf{This work reconceptualizes the evaluation of reasoning capabilities into a general and rigorous statistical testing framework.} As shown in Figure~\ref{fig:framework}, it comprises three critical components: {\it synthetic data generation, token perturbation, and statistical hypothesis testing}. 
This general framework is designed to bypass the complications of evaluation set contamination \cite{zhou2023don, ravaut2024much}, leverage insights and tools from controlled experiments, and draw statistically valid conclusions.

Our study is unique from existing work~\cite{gou2023rationality, suri2024large, mukherjee2024heuristic, wang2024will} in two folds. First,
\textbf{we are not evaluating the overall accuracy of LLMs in identifying different logical fallacies. Instead, our focus is on token bias. }
Although there are always more types of logical fallacies, we take the conjunction fallacy, syllogistic fallacy, and the "twenty-five horses" problem in graph theory as quintessential examples, which exhibit strong token biases that are more readily identifiable in their problem statements. 
\textbf{By identifying and perturbing these specific tokens, we can induce predictable shifts in LLM responses.} Second, we recognize that cognitive biases often emerge in \textit{implicit} forms in real-life scenarios, so relying on engineering fine-grained prompts ~\cite{gou2023rationality, yao2024tree, besta2024graph} to make LLMs identify specific logical fallacies is impractical for general-purpose user applications. As a result, 
we only leverage common prompting techniques that are sufficient to provide robust statistical evidence.

Comprehensive experiments on both commercial and open-sourced LLMs on large-scale synthetic datasets uncover a critical insight: it is the token bias that contributes the most to performance improvements in reasoning tasks, if any, rather than genuine advances in reasoning capabilities.

\section{The General Framework}
\label{sec: The General Framework}
Our framework is summarized in Figure~\ref{fig:framework}. This general framework is grounded on the premise that for a given reasoning task, a capable reasoning agent will consistently reach the same conclusion regardless of how the task is framed, as long as the underlying logic remains the same \citep{hastie2009rational}. This assumption lays the foundation of our null hypothesis, $\boldsymbol{H_0}$. In our setup, if an agent consistently applies reasoning in its decision-making process, the only source of failure should be the procedural mistakes during the agent's abstract reasoning steps, which we assume to come up in an i.i.d. fashion. 
Our general framework contains three major parts as follows.

\paragraph{Synthetic Data Generation} Once the underlying logic of a reasoning task is defined, we create an algorithm to generate a synthetic dataset with $n$ samples. While it is helpful to leverage LLMs for linguistic coherence in the process, the data generation should be carefully controlled, utilizing information from real-world data or established datasets to mitigate potential biases from purely AI-generated texts.
The process begins with the creation of a curated list of entities, encompassing diverse names, genders, ages, occupations, cultural backgrounds, and events where applicable, along with a textual template that dictates the structure of the task description.
By sampling from this list, we generate task descriptions that maintain the integrity and novelty of the dataset. This method ensures that while the LLM of interest might be familiar with the individual entities, it has never seen the specific combinations of these entities and narratives, thus bypassing data contamination.

The following example illustrates one approach we leverage to generate synthetic conjunction fallacy questions. We randomly sample a commonsense story curated by \citet{mostafazadeh2016corpus} and convert it into the following prompt:
\textit{Your task is to complete the last sentence of the following problem to create a conjunction fallacy quiz:}
\begin{quote}
\vspace{-1mm}
\textit{Michelle was extremely hungry. She opened the refrigerator to find nothing. Which is more likely?} \\
\textit{(a) Michelle would likely buy food at the grocery store.}\\
\textit{(b) Michelle would likely buy food at the grocery store because }
\vspace{-1mm}
\end{quote}
We expect the LLM to complete the story by providing us with a plausible reason after \textit{"because"}, such as\textit{ "she found nothing to eat at home"}. Irrespective of the LLM's completion, option (b) contains a conjunction of two events so it should always be viewed less likely.

The synthetic dataset can be dynamically generated on the fly, precluding its prior existence in any training datasets. It also allows the algorithm designers to control the dataset size, efficiently scaling their data based on the sample size required to acheive statistical validity.

\paragraph{Token Perturbation} We posit that if the LLM primarily relies on token bias, its performance on reasoning tasks will consistently improve (or degrade) as we alter some tokens in a systematic manner. This process of token perturbation generates 
$n$ matched pairs of samples, enabling us to evaluate the LLM on both the original and perturbed datasets and create a $2\times 2$ contingency table below, where $n = n_{11} + n_{12} + n_{21} + n_{22}$.

\vspace{-1mm}
\begin{table}[h]
\centering
\footnotesize
\begin{tabular}{cc|c|c}
& & \multicolumn{2}{c}{Perturbed}\\
& & Correct & Wrong \\
\cmidrule{1-4}
\multirow{2}{*}{Original}&
    Correct & $n_{11}$ & $n_{12}$ \\
    \cmidrule{2-4}
    & Wrong & $n_{21}$ & $n_{22}$
\end{tabular}
\caption{A template for the contingency table. We follow the notations in this table to define $\pi_{12}$ and $\pi_{21}$ in the next paragraph for hypothesis testing.}
\label{tab:conti}
\end{table}
\vspace{-3mm}

\paragraph{Statistical Hypothesis Testing for Matched Pairs}\label{sec:hypo}
In our context, we wish to decide whether or not some hypothesis concerning whether an agent reasons consistently is correct. The choice here lies between two decisions: accepting or rejecting the hypothesis. The decision procedure is called {\bf hypothesis testing}~\citep{lehmann1986testing}. Throughout our discussion, we use $\boldsymbol{H_0}$ to denote the null hypothesis and $\boldsymbol{H_a}$ the alternative hypothesis. 

For each of the $n$ matched pairs, let $\pi_{ab}$ denote the underlying probability of outcome $a$ for the original sample and outcome $b$ for the perturbed sample. In other words, for any nonnegative integer $m \le n$, 
\vspace{-2mm}
\begin{equation}
    \mathbb{P}(n_{ab} = m) = \binom{n}{m} \pi_{ab}^m (1-\pi_{ab})^{n-m}
\end{equation}
As $n_{ab}$ counts the
number of such pairs, $n_{ab}/n$ is the sample proportion, which is a consistent estimate of $\pi_{ab}$. The null hypothesis assumes the marginal homogeneity for binary matched pairs, i.e. $\pi_{12} = \pi_{21}$.
For small samples, we can apply an exact test conditioned on $n^* = n_{21} + n_{12}$ \cite{mosteller1952some, agresti2012categorical}. Under $\boldsymbol{H_0}, n_{21}$ follows a $\operatorname{binomial}(n^*, 1/2)$ distribution, and the corresponding $p$-value is the binomial tail probability. 
As a rule of thumb, when $n^* > 10$, the reference binomial distribution is approximately normal, and we can compute the standardized normal test statistics $z_0 = (n_{21} - n_{12})/\sqrt{n_{21} + n_{12}}$,
which is identical to the McNemar statistic \cite{mcnemar1947note}. To test the same hypotheses for a group of models, we apply the Benjamini-Hochberg Procedure~\cite{BH} to control the false discovery rate at a predetermined significance level $\alpha$.

\section{A Peek into Token Bias}\label{sec:hypo}
This section outlines the detailed hypotheses in our statistically inspired framework. We aim to determine whether LLMs are capable of genuine reasoning or whether they rely heavily on token biases.
According to \textbf{the principle of invariance} in rational decision-making~\cite{tversky1981framing, tversky1988rational}, the preferences of a rational reasoning agent should remain unaffected by the framing of equivalent decision problems. 

In a broader interpretation of invariance, we assess whether alterations in seemingly irrelevant tokens, such as name entities in problem narratives that are unrelated to the underlying logic, influence the outcomes of reasoning. A true reasoner should effectively navigate through reasoning tasks without being influenced by trivial changes in content that do not impact the fundamental logical structure. We propose a series of hypotheses, where the null hypothesis assumes the presence of a genuine reasoner. For each hypothesis, we identify specific tokens that may carry strong biases under their problem settings, and systematically alter these tokens to test their impact, while maintaining the integrity of the underlying logical structure.


\subsection{Preliminaries}\label{sec:prelim}
In this work, we integrate the conjunction fallacy and syllogistic fallacy discussed in the cognitive science literature~\citep{tversky1983extensional, kahneman2011thinking} to construct synthetic datasets on which we perform our token perturbation. This section briefly introduces the underlying logic. 
\paragraph{Conjunction Fallacy}
The most often-cited example of conjunction fallacy is called the Linda problem which is framed as follows ~\citep{tversky1983extensional}:
\textit{Linda is 31 years old, single, outspoken, and very bright. She majored in philosophy. As a student, she was deeply concerned with issues of discrimination and social justice, and also participated in antinuclear demonstrations. Which is more probable?}
\begin{quote}
\vspace{-1mm}
\textit{(a) Linda is a bank teller.}\\
\textit{(b) Linda is a bank teller and is active in the feminist movement.}
\vspace{-0.5mm}
\end{quote}
\citet{tversky1983extensional} found that humans tend to prefer option (b). However, it is logically necessary that the probability of a conjunction of two events (e.g., Linda is a bank teller, {\it and} she is active in the feminist movement) is less than the probability of either event alone. 

\paragraph{Syllogistic Fallacy} The syllogistic fallacy documents the logical failure that occurs when people are presented with syllogisms -- two premises followed by a conclusion. Ideally, if the premises are true and the logical structure is valid, the conclusion must necessarily be true. However, when the argument's structure is flawed, the conclusion may be invalid despite the surface-level logical form. Consider the following syllogism from \citet{kahneman2011thinking}:
\textit{Is this logically sound?}
\begin{quote}
\vspace{-1mm}
\textit{All roses are flowers.} \\
\textit{Some flowers fade quickly.}\\
\textit{Therefore some roses fade quickly.}
\vspace{-0.5mm}
\end{quote}

The argument is incorrect because the two premises do not imply that the set of roses and the set of flowers that fade quickly necessarily overlap. 

\subsection{Lost in Irrelevant Context}
Logical fallacies often contain misleading contexts, exploiting common cognitive biases and shortcuts in human reasoning. These fallacies can seem convincing at first glance, being effective in swaying opinions, because they resonate with intuitive yet flawed biases. For instance, conjunction fallacies present two options: one involving a single event and the other with an additional event in conjunction. This added event is particularly designed to align with the contextual background in the problem statement, leading humans or LLMs to reaffirm their preexisting beliefs. In contrast, when the additional event in the options is changed to an irrelevant one, the model is less likely to be distracted by these extraneous and irrelevant details. 

\begin{breakablealgorithm}
\caption{Genuine reasoning LLMs should withstand contextually misleading options in the problem statements.}
\begin{algorithmic}
\State \hspace{-1em} \textit{\textbf{Token perturbation: }}Assume problem $P$ is a conjunction fallacy problem with options $(a)$ and $(b)$. One option contains a event $x$ and the other contains $x$ and $y$ in conjunction. $y$ is relevant to the context of the problem statement that might mislead the LLM. In contrast, the perturbed problem $P^{\prime}$ replaces $y$ with a randomly generated $y^{\prime}$ irrelevant to the context.
\State \hspace{-1em} $\boldsymbol{H_0}$: $\pi_{12} = \pi_{21}$.
\State \hspace{-1em} $\boldsymbol{H_a}$: $\pi_{12} < \pi_{21}$. ($\pi_{12} > \pi_{21}$ is invalid.) 
\end{algorithmic}
\label{hyp:lost_in_context}
\end{breakablealgorithm}

Here is an example of such token perturbations, represented by the \textcolor{controlGroupColor}{right arrow mark}:
\textit{Kai is a community leader of Pacific Islander descent. He holds degrees in Public Administration and is passionate about preserving his cultural heritage.
Which is more probable?}
\begin{quote}
\vspace{-1mm}
\textit{(a) Kai is a law enforcement worker.} \\
\textit{(b) Kai is a law enforcement worker and \textcolor{goldenGroupColor}{participates in cultural preservation organizations} \textcolor{controlGroupColor}{$\rightarrow$ learns to play the ukulele}.}
\end{quote}

To further explain $\boldsymbol{H_0}$ and $\boldsymbol{H_a}$, the null hypothesis $\boldsymbol{H_0}$ always assumes that the LLM is a genuine reasoner and can consistently perform reasoning regardless of the superficial token changes, leading us to expect $\pi_{12} = \pi_{21}$. Meanwhile, token bias can systematically and predictably influence the LLM's performance. Accordingly, $\pi_{12} < \pi_{21}$ aligns with our expectation that misleading tokens degrade performance, but an observation of $\pi_{12} > \pi_{21}$ - where non-misleading options result in a greater decrease in performance - would be considered invalid.

\subsection{Token Bias on Widely Cited Examples in Classic Literature}
It is reasonable to suspect that most LLMs have been trained to recognize well-known logical fallacy problems. However, the question remains whether they acquire genuine reasoning skills or merely learn to falsely associate frequently appearing names - such as "Linda" in the classical Linda problem - with the correct reasoning outcomes they should have. We demonstrate an example in Figure~\ref{fig:bob} that perturbs \textcolor{goldenGroupColor}{Linda}\textcolor{controlGroupColor}{$\rightarrow$ Bob}.

\begin{breakablealgorithm}
\caption{Genuine reasoning LLM should withstand surface-level alterations to the one-shot exemplar in the problem statements.}
\begin{algorithmic}
\State \hspace{-1em} \textit{\textbf{Token perturbation: }}Assume one-shot in-context learning scenarios. $P$ has the original Linda problem as the one-shot exemplar. In contrast, the perturbed problem $P^{\prime}$ rephrases the exemplar to a persona called "Bob".
\State \hspace{-1em} $\boldsymbol{H_0}$: $\pi_{12} = \pi_{21}$.
\State \hspace{-1em} $\boldsymbol{H_a}$: $\pi_{12} > \pi_{21}$. ($\pi_{12} < \pi_{21}$ is invalid.) 
\end{algorithmic}
\label{hyp:linda_bob}
\end{breakablealgorithm}

\subsection{Token Bias on Well-Known Entity Names}
Celebrity names inherently carry a rich contextual background that LLMs learn from massive training data. We hypothesize that by replacing a celebrity name with a generic one in a conjunction fallacy problem, thereby dissociating the link to this contextual backdrop, we might see performance improvements in LLMs, and such results would underscore the potential deficiency in their genuine reasoning capabilities. 

\begin{breakablealgorithm}
\caption{Genuine reasoning LLMs should withstand irrelevant alterations to name entities in problem statements}
\begin{algorithmic}
\State \hspace{-1em} \textit{\textbf{Token perturbation: }}Assume $P$ is a conjunction fallacy problem that involves a celebrity. In contrast, the perturbed problem $P^{\prime}$ replaces the celebrity name with a generic one.

\State \hspace{-1em} $\boldsymbol{H_0}$: $\pi_{12} = \pi_{21}$.
\State \hspace{-1em} $\boldsymbol{H_a}$: $\pi_{12} < \pi_{21}$. ($\pi_{12} > \pi_{21}$ is invalid.)

\State \hspace{-1em} \textit{\textbf{Token perturbation: }}Assume $P$ is a mathematical reasoning problem that involves a classic entity name in its story narratives. In contrast, the perturbed problem $P^{\prime}$ alters the classic name to another random one.

\State \hspace{-1em} $\boldsymbol{H_0}$: $\pi_{12} = \pi_{21}$.
\State \hspace{-1em} $\boldsymbol{H_a}$: $\pi_{12} > \pi_{21}$. ($\pi_{12} < \pi_{21}$ is invalid.)
\end{algorithmic}
\label{hyp:celebrity}
\end{breakablealgorithm}

Here is an example of the token perturbation involving celebrity names:
\textit{\textcolor{goldenGroupColor}{Taylor Swift}\textcolor{controlGroupColor}{$\rightarrow$ Lauren} will embark on another tour in 2027. Which outcome do you think is more likely?}
\begin{quote}
\vspace{-1mm}
\textit{(a) Her first show is a flop.} \\
\textit{(b) Her first show is a flop but she will eventually sell over a million tickets for the entire tour.}
\end{quote}

Here is another example of the token perturbation applied to the classic "twenty-five horses" problem in mathematical reasoning, as referenced in Figure~\ref{fig:horses}. Note that perturbing the numbers is optional, but the total number of animals must always be a square number: \textit{There are \textcolor{goldenGroupColor}{twenty-five}\textcolor{controlGroupColor}{$\rightarrow$ thirty-six} \textcolor{goldenGroupColor}{horses}\textcolor{controlGroupColor}{$\rightarrow$ bunnies} among which you need to find out the fastest three. You can conduct a race among at most \textcolor{goldenGroupColor}{five}\textcolor{controlGroupColor}{$\rightarrow$ six} to find out their relative speed. At no point, you can find out the actual speed of the \textcolor{goldenGroupColor}{horse}\textcolor{controlGroupColor}{$\rightarrow$ bunnies} in a race. Find out the minimum number of races which are required to get the top \textcolor{goldenGroupColor}{five}\textcolor{controlGroupColor}{$\rightarrow$ six} \textcolor{goldenGroupColor}{horses}\textcolor{controlGroupColor}{$\rightarrow$ bunnies}.}

\subsection{Token Bias in Reasoning about Sets}
The syllogistic fallacy involves reasoning about sets, utilizing specific quantifiers such as "all" and "some" to specify the distribution of variables. Our investigation centers on whether LLMs overfit to these tokens of quantifiers, relying heavily on their presence to generate answers that appear correct. By rephrasing these tokens with other words that convey the same meaning, we can test the robustness of LLMs' reasoning abilities.

\begin{breakablealgorithm}
\caption{Genuine reasoning LLM should withstand irrelevant alterations to the quantifiers in problem statements.}
\begin{algorithmic}
\State \hspace{-1em} \textit{\textbf{Token perturbation: }}Assume $P$ is a syllogistic fallacy problem with quantifier tokens like "All" and "Some". In contrast, the perturbed problem removes "All" or rephrases "all" and "some" to different words with the same meaning.
\State \hspace{-1em} $\boldsymbol{H_0}$: $\pi_{12} = \pi_{21}$.
\State \hspace{-1em} $\boldsymbol{H_a}$: $\pi_{12} > \pi_{21}$. ($\pi_{12} < \pi_{21}$ is invalid.)
\end{algorithmic}
\label{hyp:sets}
\end{breakablealgorithm}

Here is an example of such token perturbations:
\textit{Is it logically sound? \textcolor{goldenGroupColor}{All roses}\textcolor{controlGroupColor}{$\rightarrow$ Roses} are flowers. \textcolor{goldenGroupColor}{Some}\textcolor{controlGroupColor}{$\rightarrow$ A subset of} flowers fade quickly. Therefore, \textcolor{goldenGroupColor}{some}\textcolor{controlGroupColor}{$\rightarrow$ A subset of} roses fade quickly. }

Continuing with the exploration of token bias in syllogistic fallacies, we propose an intriguing rephrasing of the syllogism's narrative by incorporating the names of reputable news agencies and universities. While adding the tokens of their names does not alter the logic, it could influence how LLMs perceive and process the information. LLMs prone to token bias might erroneously increase their confidence in the trustworthiness and credibility of the stories, based purely on the association with these respected institutions. 

\begin{breakablealgorithm}
\caption{Genuine reasoning LLM should withstand alterations to the narrative.}
\begin{algorithmic}
\State \hspace{-1em} \textit{\textbf{Token perturbation: }}Assume $P$ is the original problem. The perturbed problem $P^{\prime}$ adds or modifies specific tokens in the problem statement to reframe its narratives without changing the logic structure.
\State \hspace{-1em} $\boldsymbol{H_0}$: $\pi_{12} = \pi_{21}$.
\State \hspace{-1em} $\boldsymbol{H_a}$: $\pi_{12} < \pi_{21}$ or $\pi_{12} > \pi_{21}$.
\end{algorithmic}
\label{hyp:framing}
\end{breakablealgorithm}

To remove potential token bias from the pattern \textit{"All..., Some..., Some..."}, we regard perturbed problems $P^{\prime}$ in Hypothesis~\ref{hyp:sets} as the original problems $P$ here, as shown in the example below:
\textit{Is it logically sound? \textcolor{goldenGroupColor}{Roses}\textcolor{controlGroupColor}{$\rightarrow$ In a recent publication by Bloomberg, it was noted that roses} are flowers. \textcolor{goldenGroupColor}{A subset of}\textcolor{controlGroupColor}{$\rightarrow$ Research from MIT supports the finding that a subset of} flowers fade quickly. Therefore, a subset of roses fade quickly. }

To ensure a more comprehensive comparison, we also alter tokens to satirical sources like The Onion and less reputable institutions, noting that these names never impact the logical structure of the problems. Here is an example: \textit{Is it logically sound? \textcolor{goldenGroupColor}{Roses}\textcolor{controlGroupColor}{$\rightarrow$ In a recent publication by the Daily Rumor, it was noted that roses} are flowers. \textcolor{goldenGroupColor}{A subset of}\textcolor{controlGroupColor}{$\rightarrow$ An anonymous blog post writes the finding that a subset of} flowers fade quickly. Therefore, a subset of roses fade quickly. }

\subsection{Leaking Hint Tokens}
Just as a proficient student doesn't need hints to excel in a math exam, a reasoning agent should solve logical problems effectively without explicit cues. Besides, even if a student answers all problems correctly but the examlet provides all the reasoning steps, we may still question whether the student really understands the reasoning. Our experiments deliberately leak important hints that we expect a genuine reasoner to figure out itself in its intermediate reasoning steps. 

\begin{breakablealgorithm}
\caption{Genuine Reasoning LLMs should not rely on hint tokens to derive correct inferences.}
\begin{algorithmic}
\State \hspace{-1em} \textit{\textbf{Token perturbation: }}Assume in-context learning scenarios. The perturbed problem $P^{\prime}$ explicitly adds hint tokens in its prompts, such as the name of the logical fallacy or detailed guidance on the correct reasoning, while $P$ does not.
\State \hspace{-1em} $\boldsymbol{H_0}$: $\pi_{12} = \pi_{21}$.
\State \hspace{-1em} $\boldsymbol{H_a}$: $\pi_{12} < \pi_{21}$. ($\pi_{12} > \pi_{21}$ is invalid.) 
\end{algorithmic}
\label{hyp:hint}
\end{breakablealgorithm}

Here is an example of such token perturbations:
\textit{Marsha Ellis, 42, is an African American transgender female. She is an ardent advocate for gender-affirming rights and environmental protection. 
Which is more probable?\textcolor{controlGroupColor}{$\rightarrow$ Please be aware
that this is a problem on the conjunction fallacy.}}
\begin{quote}
\vspace{-1mm}
\textit{(a) Marsha is a research scientist.} \\
\textit{(b) Marsha is a research scientist and volunteers at LGBTQ+ health centers.}
\end{quote}

We also manually craft a detailed chain-of-thought instructions~\cite{wei2022chain} that teach LLMs about correct reasoning steps and potential logical pitfalls, as shown in Appendix~\ref{app:prompt}.

\section{Experiment}
Our experiments aim to rigorously test the reasoning capabilities of LLMs through the hypotheses in Section~\ref{sec:hypo} on token bias. More comprehensive results are included in Appendix~\ref{app:full_results}.
In all experiments, we run $n$ trials for each "model-prompting method" pair, depending on how many synthetic data samples are related to each hypothesis, and then perform a McNemar test. We apply the Benjamini-Hochberg Procedure and reject the null hypothesis if the p-value is less than $\alpha = 0.05$.

\subsection{Models}
We experiment with a variety of commercial and open-sourced LLMs for a thorough study, including OpenAI gpt-3.5-turbo, gpt-4-turbo, and gpt-4o~\cite{brown2020language, achiam2023gpt}, Meta llama-3-70b-instruct, llama-3-8b-instruct, and llama-2-70b-chat~\cite{touvron2023llama}, Anthropic claude-3-opus-20240229 and claude-3-sonnet-20240229~\cite{anthropic2024claude}, and Mistral mistral-large-largest~\cite{jiang2023mistral}.

\subsection{Synthetic Dataset Generation}
\label{sec: data generation}

We leverage data sources such as occupational statistics \citep{USDL2024}, commonsense stories \citep{mostafazadeh2016corpus}, CNN news stories \citep{see2017get}, common disease symptom pairs \cite{kaggle2024diseasesymptom}, celebrity names~\cite{rosenberg2021, forbes2024celebrity100wiki}, objects vocabularies \cite{eslkidsvocabulary} and common U.S. news media \cite{wikipedia2024newsmediausa, pew2011top25} to curate lists of entities to generate synthetic data. We outline our well-controlled data generation process and the samples used for each hypothesis in Appendix~\ref{app:synthetic_data}.

\begin{figure*}
  \centering
  \begin{subfigure}{\textwidth}
  \vspace{-1mm}
    \includegraphics[width=\linewidth]{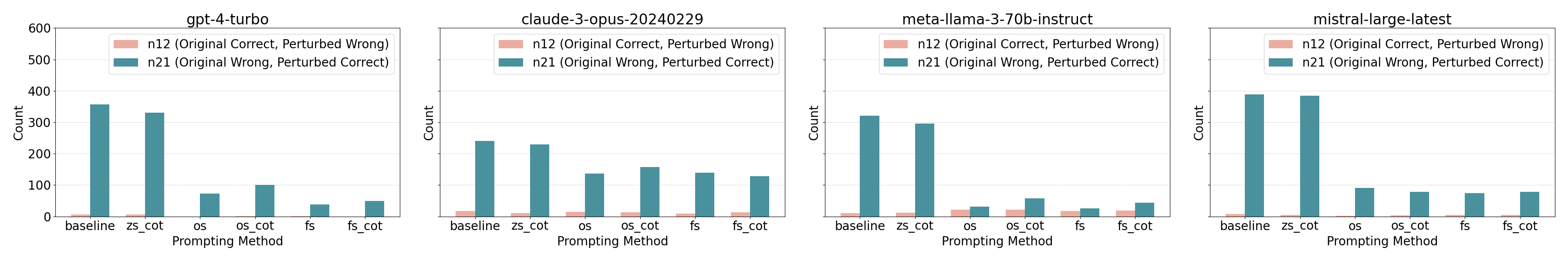}
    \caption{Experimental results for Hypothesis~\ref{hyp:lost_in_context} ($n = 400$). The perturbed problems alternate options contextually relevant to the problem statements to irrelevant ones. We run all different prompt methods. To reject the null, we expect $\textcolor{goldenGroupColor}{n12} < \textcolor{controlGroupColor}{n21}.$ We conclude that LLMs fail to reason against contextually misleading options in conjunction fallacy problems.}
    \label{fig:h1}
  \end{subfigure}
  \vspace{-4mm} 

  \begin{subfigure}{\textwidth}
    \includegraphics[width=\linewidth]{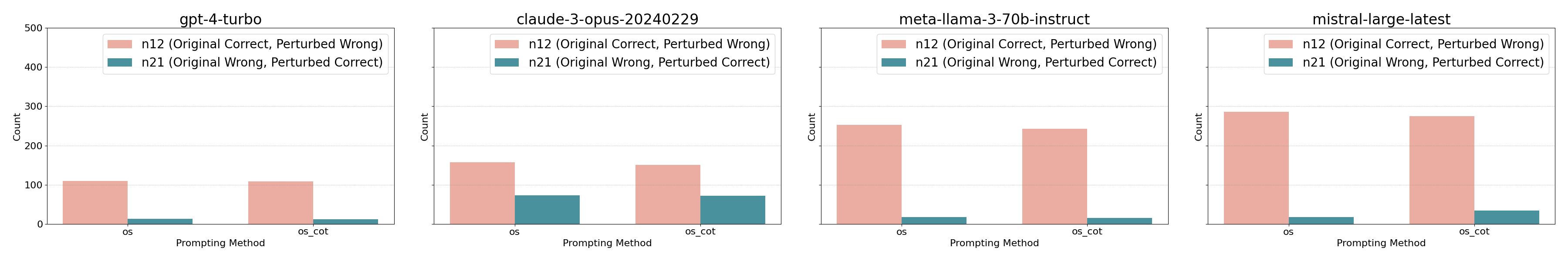}
    \caption{Experimental results for Hypothesis~\ref{hyp:linda_bob} ($n = 500$). The perturbed problems alternate the name classic "Linda" to "Bob" in in-context learning exemplars. We run one-shot with and without chain-of-thought prompts. To reject the null, we expect $\textcolor{goldenGroupColor}{n12} > \textcolor{controlGroupColor}{n21}$. We conclude that LLMs possess strong token bias to the name "Linda" frequently appearing in classic literature.}
    \label{fig:h2}
  \end{subfigure}
  \vspace{-4mm} 

  \begin{subfigure}{\textwidth}
    \includegraphics[width=\linewidth]{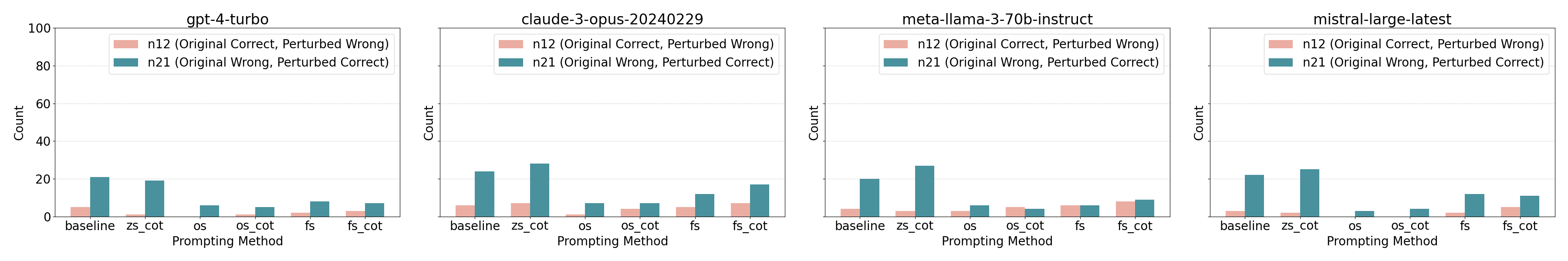}
    \caption{Experimental results for Hypothesis with celebrity names~\ref{hyp:celebrity} ($n = 100$). The perturbed problems alternate the celebrity name to a generic one in problem statements. We run all different prompt methods. To reject the null, we expect $\textcolor{goldenGroupColor}{n12} < \textcolor{controlGroupColor}{n21}.$ We conclude that LLMs are frequently misled by celebrity names in problem statements that are irrelevant to logical essence.}
    \label{fig:h3}
  \end{subfigure}
  \vspace{-4mm} 

  \begin{subfigure}{\textwidth}
    \includegraphics[width=\linewidth]{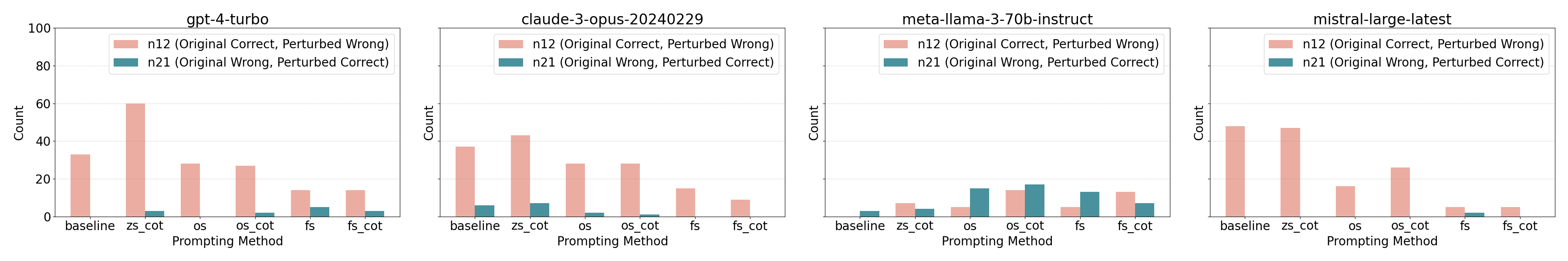}
    \caption{Experimental results for Hypothesis~\ref{hyp:sets} ($n = 200$). The perturbed problems alternate tokens "All" and "Some" to different but equivalent expressions in syllogisms. We run all different prompt methods. To reject the null, we expect $\textcolor{goldenGroupColor}{n12} > \textcolor{controlGroupColor}{n21}.$ We conclude that most LLMs rely on patterns \textit{"All..., Some..., Some..."} for reasoning about syllogism.}
    \label{fig:h4}
  \end{subfigure}
  \vspace{-4mm} 

  \begin{subfigure}{\textwidth}
    \includegraphics[width=\linewidth]{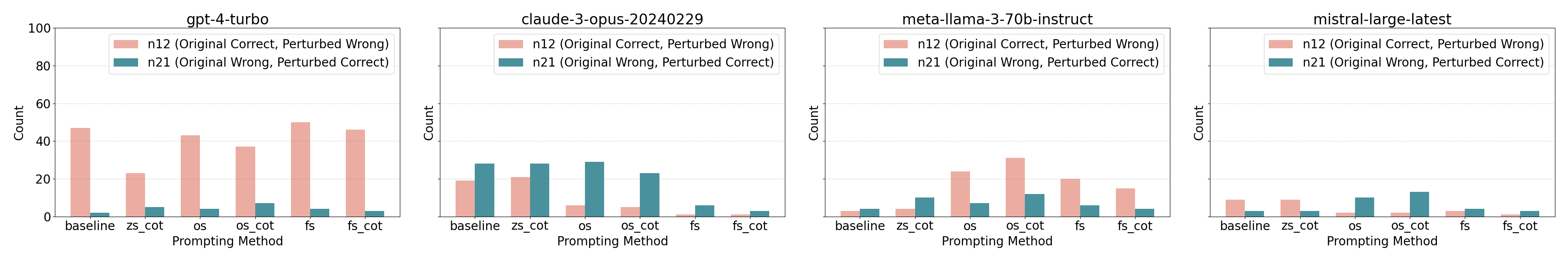}
    \caption{Experimental results for Hypothesis~\ref{hyp:framing} ($n = 200$). The perturbed problems add the names of trustworthy news agencies and universities to alter the narratives of syllogisms. We run all different prompt methods. To reject the null, we expect $\textcolor{goldenGroupColor}{n12} > \textcolor{controlGroupColor}{n21}.$ We conclude that LLMs might be misled by reputable names irrelevant to the logical structure.}
    \label{fig:h5}
  \end{subfigure}
  \vspace{-4mm} 

  \begin{subfigure}{\textwidth}
    \includegraphics[width=\linewidth]{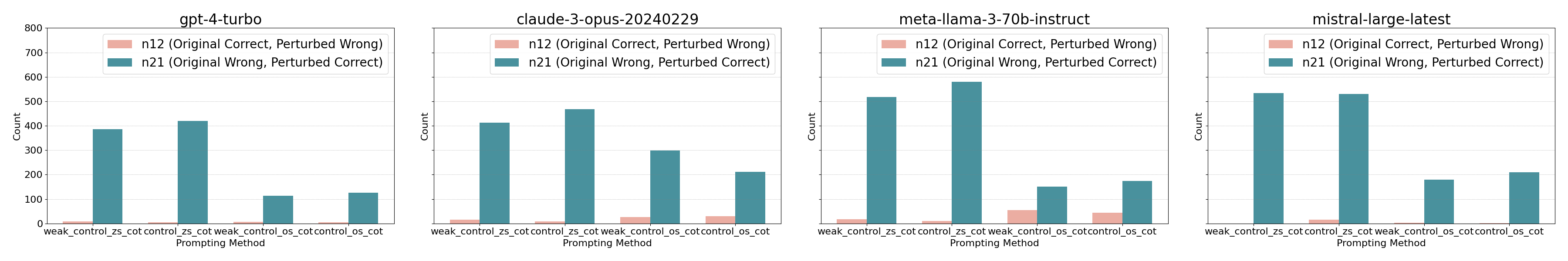}
    \caption{Experimental results for Hypothesis~\ref{hyp:hint} ($n = 800$). The perturbed problems leak hint tokens, either weak or strong hints in problem statements. We run zero-shot and one-shot prompt methods. To reject the null, we expect $\textcolor{goldenGroupColor}{n12} < \textcolor{controlGroupColor}{n21}.$ We conclude that LLMs still heavily rely on hint tokens for solving logical fallacy problems well.}
    \label{fig:h6}
  \end{subfigure}
  \vspace{-6mm} 

  \caption{Our controlled experiments cast doubt on the genuine reasoning capabilities of LLMs. In this figure, each pair of \textcolor{goldenGroupColor}{histo}\textcolor{controlGroupColor}{grams} stuck together represents a comparison in the contingency table~\ref{tab:conti} for McNemar's Tests.} 
  \label{fig:all_results} 
\end{figure*}

\subsection{Prompting Methods}
We implemented commonly used prompting strategies that are sufficient for evaluating the null hypotheses within our framework. The specific prompting techniques we utilized are as follows, with their corresponding notations presented in Figure~\ref{fig:all_results}: \textit{Baseline}: Directly answering the question without additional instructions. \textit{Zero-shot chain-of-thought (zs\_cot)}: Includes the instruction "Let us think step by step"~\cite{wei2022chain}. \textit{One-shot (os)}: Involves a single in-context learning example~\cite{brown2020language}. \textit{Few-shots (fs)}: Utilizes three in-context examples. Similarly, we have \textit{os\_cot} and \textit{fs\_cot}. We also include \textit{weak\_control\_zs/os\_cot} and \textit{control\_zs/os\_cot} for Hypothesis~\ref{hyp:hint} representing prompts with additional weak or strong hints, as detailed in Appendix~\ref{app:prompt}.

\subsection{Hypothesis Testing Results}
\paragraph{Testing of Hypothesis 1: LLMs Would Fail at Misleading Options}
We evaluate LLMs on all conjunction fallacy problems with misleading options ($n$=400). Figure~\ref{fig:h1} and Table~\ref{table:h1} show a significant decline in success rate when contextually misleading options in conjunction fallacy problems are replaced with random alternatives. 
The random ones are no longer relevant to the problem statements, so all LLMs become less likely to be swayed by background information that is not logically important. We, therefore, reject almost all null hypotheses.

\paragraph{Testing of Hypothesis 2: LLMs Would Fail Due to Surface Level Change in the Exemplar}
We evaluate LLMs under in-context learning scenarios for solving conjunction fallacies ($n$=500). Figure~\ref{fig:h2} and Table~\ref{table:h2} show consistent performance drop on all LLMs when the name "Linda," frequently used in classic reasoning tasks, is substituted with "Bob" in one-shot exemplars. Such a change should not influence outcomes for genuine reasoners, as the specific name used is irrelevant to the logical process.

\paragraph{Testing of Hypothesis 3: LLMs Would be Misled by Celebrity Names}
We evaluate LLMs on variants of conjunction fallacies that contain a celebrity name ($n$=100). We observe in Figure~\ref{fig:h3} and Table~\ref{table:h3} to Figure~\ref{fig:h1} that celebrity names appeared in problem statements frequently mislead LLMs into the celebrity's background, which is not helpful in solving logical fallacy problems but reduces accuracy, leading us to reject all null hypotheses.

\paragraph{Testing of Hypothesis 4: LLMs Would Fail at Synonyms of Classic Quantifiers}
We assess LLMs on syllogisms ($n$=200). Figure~\ref{fig:h4} and Table~\ref{table:h4} reveal that, in most instances, we should reject the null hypotheses, except for llama-3-70b-instruct. Most LLMs demonstrate insufficient robustness when patterns \textit{"All..., Some..., Some..."} commonly used in classic syllogistic fallacy problems are substituted with synonyms.

\paragraph{Testing of Hypothesis 5: LLMs Would Be Misled by Names Linked to Reputable Entities}
We evaluated the impact of names linked to reputable data sources in syllogisms ($n$=200). Figure~\ref{fig:h5} and Table~\ref{table:h5} demonstrate that some LLMs are indeed misled by the inclusion of these authoritative names, especially GPT-4 and LLaMA-3-70B. Generally, LLMs tend to falsely believe that these narratives are more trustworthy and, thereby, ignore the logical fallacy in them. As a result, we reject about half of the null hypotheses. Results from using the names of less credible sources are shown in Appendix~\ref{table:h5_framing} for comparison.

\paragraph{Testing of Hypothesis 6: LLMs Still Heavily Rely on Hint Tokens}
We evaluated the performance of LLMs with and without the presence of hints ($n$=200). Figure~\ref{fig:h6} and Table~\ref{table:h6} indicate that LLMs still heavily rely on hints to achieve ideal performance, so we reject all the null hypotheses.

\section{Related Work}


\paragraph{The Reasoning Capabilities of LLMs}
There is an increasing number of works aim to improving the reasoning performance of LLMs~\cite{ouyang2022training, zhou2022learning, lyu2023faithful, hao2023reasoning, xu2024can, putta2024agent, kumar2024training, yao2024tree, cai2024t}, evaluating and critiquing their reasoning abilities~\cite{zhou2020temporal, hong2023closer, huang2023large, shi2023large, sprague2024cot, turpin2024language, xiao2024logicvista}, trying to understand their reasoning processes~\cite{wei2022emergent, merrill2023expresssive, ma2023training, lanham2023measuring, merrill2024logic, yang2024large, chen2024premise, jin2024impact}, as well as related surveys~\cite{qiao2022reasoning, huang2022towards, ahn2024large, giadikiaroglou2024puzzle, liang2024internal, zheng2024attention} that comprehensively explore the reasoning capabilities of LLMs. Our work aligns with efforts that critique the generalization of LLMs' reasoning capabilities, moving beyond accuracy-based benchmarks~\cite{talmor2018commonsenseqa, cobbe2021training, srivastava2022beyond, fu2023chain, yang2018hotpotqa}, \textbf{which primarily focus on overall question-answering accuracy, often without considering question augmentations or variations}. Instead, we step one-level deeper into these problem statements, \textbf{investigating potential token biases that may create the illusion of improved logical reasoning performance, while remaining vulnerable to irrelevant token perturbations.} By examining these subtleties, our approach provides a more nuanced understanding of the limitations in LLM's reasoning capabilities. Besides, we reformulate the evaluation as hypothesis testing, providing results with statistical guarantees.

\paragraph{Cognitive Biases and Logical Fallacies in LLMs}
Recent studies~\cite{gardner2020evaluating, hagendorff2023human, lin2023mind, talboy2023challenging, binz2023using, ullman2023large, mitchell2023debate} analyze the biases in LLMs with synthetic datasets. For example, \citet{tamkin2023evaluating} uses an LLM to generate prompts that
reveals patterns of discriminations in LLMs.
\citet{echterhoff2024cognitive} proposes a set of LLM-simulated experiments in a specific context. Although existing works~\cite{mukherjee2024heuristic, macmillan2024ir, wang2024will, suri2024large, jin2022logical} study more kinds of fallacy types in human psychology, they approach problems at a coarse level and only emphasize accuracy. Our study goes into a more fine-grained level with a series of hypotheses. We provide statistical guarantees and quantitative analyses of token bias that can be carefully and systematically tuned. Besides, \citet{gou2023rationality} presents the Rationality of Thought (RoT), decomposing responses into six predefined steps with hand-crafted prompt engineerings. Our work focuses on general prompting strategies that are sufficient to validate or reject our hypotheses.

\section{Discussion}
This work reconceptualizes the evaluation of the reasoning behavior of LLMs through the lens of token bias. 
The statistical evidence presented in our hypothesis-testing framework contributes to the larger discussion that LLMs do not always apply reasoning consistently in their decision-making processes. Instead, they primarily rely on token bias for response generation. This suggests that chain-of-thought prompting~\cite{wei2022chain, wang2022towards} or in-context learning~\cite{brown2020language, min2022rethinking, lyu2022z, wang2022towards} may not elicit actual reasoning but instead result in semantic shortcuts for LLMs to imitate the desired behavior at superficial levels. 
These findings raise concerns about the extent to which LLMs truly engage in reasoning. Further investigations are needed to uncover the underlying mechanisms and limitations of LLMs' reasoning capabilities.

\section{Limitations}
This hypothesis-testing framework is specifically designed for multiple-choice or yes/no questions and is not applicable to open-ended responses. It relies on LLMs with strong instruction-following capabilities to consistently produce responses that include the selected options, though we find that LLMs can generally follow these instructions in most cases. In addition, smaller LLMs, such as llama-3-8b-instruct, with lower instruction following capabilities may contain more confounders besides the token bias, which could weaken our hypothesis testing results. As a result, we mainly focus on state-of-the-art LLMs. Moreover, the finding of token biases requires manual efforts. We also acknowledge that there are likely other hypotheses and assumptions that a genuine reasoner should satisfy. Our current study focuses on the conjunction fallacy, syllogistic fallacy, the "twenty-five horses" problem in graph theory, and their variants to demonstrate our framework. These are quintessential examples and the framework could include a broader range of hypotheses, fallacy types, data modalities, and reasoning tasks.

\section{Acknowledgement}
This work was supported by NSF grant CCF-2112665 (TILOS), which provides funding for B.J. and C.J. It was also funded in part by the Laboratory Directed Research and Development (LDRD) program at Argonne National Laboratory, with support from the Office of Science, U.S. Department of Energy, under Contract No. DEAC02-06CH11357. Y. X. and W. J. Su acknowledge support from the NSF HDR TRIPODS award (CCF-1934876). The authors thank John K. Hutchison for his valuable suggestions regarding the motivation of this study.

\bibliography{custom}

\newpage
\appendix
\onecolumn

\section{The Original Linda Problem in \citet{tversky1983extensional}}
\label{app:linda}

The original Linda problem is framed as follows ~\citep{tversky1983extensional}:
\begin{quote}

Linda is 31 years old, single, outspoken, and very bright. She majored in philosophy. As a student, she was deeply concerned with issues of discrimination and social justice, and also participated in antinuclear demonstrations. Which is more probable?

\begin{enumerate}
    \item Linda is a bank teller.
    \item Linda is a bank teller and is active in the feminist movement.
\end{enumerate}

\end{quote}

Here is an example of GPT-4o explaining the Linda Problem: \url{https://chatgpt.com/share/eff10b9d-d219-4806-9cb9-d2d9104c0e83}.

Our ``Bob'' version of this problem is as follows:
\begin{quote}
Bob is 29 years old, deeply passionate about environmental conservation, and volunteers his weekends at local park clean-ups. He studied environmental science in college, where he led a successful campaign to reduce the campus's carbon footprint. Bob is also an avid cyclist and promotes sustainable living practices whenever possible. Based on this information, which is more possible?
\begin{enumerate}
    \item Bob works for a renewable energy company and is an active member of a local environmental advocacy group.
    \item Bob works for a renewable energy company.
\end{enumerate}
\end{quote}

\section{Prompts in Hypothesis~\ref{hyp:hint}}\label{app:prompt}
This section includes the detailed prompts we use to evaluate the influences from weak and strong hints. These prompts are added to either the zero-shot chain-of-thought or the one-shot chain-of-thought prompts.

\subsection{Weak Hint} 
\paragraph{For Problems on Conjunction Fallacies} Your task is to answer the following question by explicitly selecting either option (a), (b), etc. Please be aware that this is a Linda Problem designed to explore the concept of the conjunction fallacy. Here is the question and let’s think step by step.

\paragraph{For Problems on Syllogistic Fallacies}
Your task is to answer the following question by explicitly saying 'Yes' or 'No'. Please be aware that this is a Linda Problem designed to explore the concept of the syllogistic fallacy.

\subsection{Strong Hint} 
\paragraph{For Problems on Conjunction Fallacies}Your task is to answer the following question by explicitly selecting either option (a), (b), etc. Please aware that this is a Linda Problem designed to explore the concept of the conjunction fallacy. The conjunction fallacy occurs when individuals incorrectly judge the conjunction of two events as more probable than one of the events alone. For instance, many might believe that Linda, who is described as a bright, single woman deeply concerned with discrimination and social justice, is more likely to be both a bank teller and active in the feminist movement than just a bank teller. This judgment violates the basic probability rule: the probability of a conjunction, P(A and B), is always less than or equal to the probabilities of its constituents, P(A) or P(B). This error often stems from the representativeness heuristic, where people estimate the likelihood of an event by how closely it matches their mental prototype. To correctly solve problems like this, you must adopt probabilistic thinking: abstract the problem from its narrative context and focus solely on the probabilistic models. Ignore all extraneous background information and consistently choose the option involving a single event as it statistically holds a higher likelihood than the conjunction of multiple events. Here is the question and let’s think step by step.

\paragraph{For Problems on Syllogistic Fallacies} Your task is to answer the following question by explicitly saying 'Yes' or 'No'.
Please aware that this is a Syllogistic Fallacy Problem. This type of reasoning is known as a syllogism. 
Pay close attention to quantifiers such as 'All', 'Some', 'No', or similar terms. These terms help define the distribution of properties or elements within the given groups or categories in the premises. 
Next, assess whether the attribute ascribed in the conclusion necessarily follows from the attributes described in the premises. 
Consider if the subset described in the second premise encompasses or overlaps with the elements in the first premise that are carried into the conclusion. 
A common pitfall in syllogistic reasoning is the erroneous assumption that a characteristic of a subset of a group (from the premises) applies to another 
subset of the same or different group (in the conclusion), without explicit justification. Ignore the background information about the objects and focus on the logical structure of the argument.
Here is an example.

\section{Synthetic Data Generation}\label{app:synthetic_data}
In this section, we outline the controlled synthetic data generation process. For each variant, we generate 100 synthetic data samples.

\subsection{Conjunction Fallacy}
We create several variants of the conjunction fallacy problem discussed in the original work by \citet{tversky1983extensional}:

\paragraph{Variant 1} The original Linda Problem. We maintain the narrative structure of the original Linda Problem described in Appendix~\ref{app:linda}. We ask GPT-4 to randomly pick reasonable personal details such as name, race, gender identity, age, and major, forming a short biography. GPT-4 then crafts two options $(a)$ and $(b)$ for each problem, both of which contain the same randomly selected occupation from \citet{USDL2024} like \textit{"Linda is a bank teller"}. The longer option also contains a hobby that must be relevant to the bio like \textit{"active in the feminist movement"}. 

The prompt used to generate the bio is as follows, where \texttt{\{random\_gender\}, \{random\_race\},\{random\_age\}} are sampled from a pre-defined random function:

\begin{quote}
\texttt{Your task is to write a short bio for a random person within 100 words. You shall pick a random name, use gender \{random\_gender\}, race \{random\_race\}, and an age \{random\_age\}. The bio should describe the college majors, some personal characters, and interests. Keep the bio short. For example, 'Linda is 31 years old, single, outspoken, and very bright. She majored in philosophy. As a student, she was deeply concerned with issues of discrimination and social justice, and also participated in anti-nuclear demonstrations. Write another example here:}
\end{quote}

We then follow-up the conversation with the following prompt:
\begin{quote}
\texttt{Your next step is to find a hobby or activity that the person mentioned before will be interested in based on your experience. 
The hobby or activity must be relevant to the bio descriptions. 
In the example above, we can say that 'Linda is active in the feminist movement.' because her bio says she was concerned with discrimination and social justice. 
Please keep your answer in one sentence and begin with that person's name, but refrain from using any words used in the bio.
}
\end{quote}

To create token bias, we generate a random hobby using the following:
\begin{quote}
    \texttt{Your task is to find a random hobby or activity, and keep your answer short in one sentence. For example, you can say 'cook Asian foods.'}
\end{quote}

\paragraph{Variant 2} In the original paper, the following variant of the conjunction fallacy problem is also presented:

\begin{quote}
    John P. is a meek man, 42 years old, married with two children. His neighbors describe him as mild-mannered but somewhat secretive. He owns an import-export company based in New York City, and he travels frequently to Europe and the Far East. Mr. P. was convicted once for smuggling precious stones and metals (including uranium) and received a suspended sentence of 6 months in jail and a large fine. Mr. P. is currently under police investigation. Which one is more likely?
\begin{enumerate}
    \item Mr. P. killed one of his employees.
    \item Mr. P. killed one of his employees to prevent him from talking to the police.
\end{enumerate}
\end{quote}

The conjunction of two events in the second option is connected by the word `to.' To create this dataset, we sample a random story from the collection of commonsense stories \citep{mostafazadeh2016corpus} and CNN news stories \citep{see2017get}. We use all the sentences in the story as the context of the conjunction fallacy problem except for the last one. We use the last sentence as the first option in the problem. As for the second option, we append the last sentence, add the connecting word `to,' and then we prompt GPT-4 to complete the second option. The prompt used here is similar to that discussed in section~\ref{sec: The General Framework}. 

To perform token perturbation, we further prompt GPT-4 with the following:

\begin{quote}
    \texttt{Your next task is to complete the last sentence of the same problem but make sure your completion after 'to' is now irrelevant to the content intentionally:}
\end{quote}

\paragraph{Variant 3} This is the same as the last variant, except that we use `because' as the connecting word.

\paragraph{Variant 4} This is the same as the last variant, except that we use `so that' as the connecting word.

\paragraph{Variant 5}
In the original paper, the following variant of the conjunction fallacy is discussed:
\begin{quote}
A 55-year-old woman had pulmonary embolism documented angiographically 10 days after a cholecystectomy. Which is more likely?
\begin{enumerate}
    \item dyspnea and hemiparesis
    \item hemiparesis
\end{enumerate}
\end{quote}

Inspired by this example, we randomly sample a disease and its corresponding symptoms from \cite{kaggle2024diseasesymptom} and apply the following prompt to generate a conjunction fallacy problem:

\begin{quote}
\texttt{Your task is to create another conjunction fallacy quiz following the format in the example below. Do not mention the name 'conjunction fallacy.' 
You should pick a random name for the patient, use gender \{random\_gender\} race \{random\_race\}, an age \{random\_age\} and the disease \{random\_disease\} in your new problem statement. The question should be 'Which one is more likely?' followed by two options (a) and (b), one of which should be a subset of the other. You can randomly switch the order of which option is (a) and which is (b). You should use the symptoms \{random\_symptom\_one\} in both options and add \{random\_symptom\_two\} to the longer option only. Do not make any changes to the given disease or the symptoms. Here is the new problem:}
\end{quote}

We then prompt GPT-4 for an irrelevant symptom:

\begin{quote}
\texttt{Your task is to create another conjunction fallacy quiz following the format in the example below. Do not mention the name 'conjunction fallacy.' You should pick a random name for the patient, use gender \{random\_gender\} race \{random\_race\}, an age \{random\_age\} and the disease \{random\_disease\} in your new problem statement. The question should be 'Which one is more likely?' followed by two options (a) and (b), one of which should be a subset of the other. You can randomly switch the order of which option is (a) and which is (b). You should use the symptoms \{random\_symptom\_one\} in both options. You should add another random symptoms to the longer option only, which must be completely irrelevant to the disease \{random\_disease\} intentionally. Do not make any changes to the given disease or the symptoms. Here is the new problem:}
\end{quote}

\paragraph{Variant 6}

In the original paper, the following variant of the conjunction fallacy is discussed:
\begin{quote}
Suppose Bjorn Borg reaches the Wimbledon finals in 1981. Which is more likely?

\begin{enumerate}
    \item Borg will lose the first set
    \item Borg will lose the first set but win the match
\end{enumerate}
\end{quote}

Inspired by this example, we randomly sample celebrity names from the Times Person of the Year~\cite{rosenberg2021} and Forbes Celebrity 100~\cite{forbes2024celebrity100wiki} and apply the following few-shot prompt to generate a conjunction fallacy problem.  

\begin{quote}
\texttt{Create one example that look like this:\\
Suppose [celebrity is going to do something]. Which is more likely:\\
(a) [Something unlikely for this person]\\
(b) [Something unlikely for this person] but [something extremely likely for this person]\\\\
Here are some examples:\\\\
Suppose Taylor Swift is going to have another tour in 2027. Which is more likely:\\
(a) Her first show is a flop.\\
(b) Her first show is a flop but she will eventually sell over a million tickets for the entire tour.\\
Suppose Barack Obama is running for president in 2024. Which is more likely:\\
(a) Barack Obama will win the national popular vote\\
(b) Barack Obama will win the national popular vote but lose the Electoral College vote\\
Suppose Bjorn Borg reaches the Wimbledon finals. Which outcome is more likely?\\
(a) Borg will lose the first set\\
(b) Borg will lose the first set but win the match\\
Complete the following. Do not output anything else.\\
Suppose \{random\_celebrity\}}
\end{quote}

For Hypothesis~\ref{hyp:lost_in_context}, we include Variant 2,3,4 and 5, resulting in $n = 400$ samples. For Hypothesis~\ref{hyp:linda_bob}, we include Variant 2,3,4,5 and 6, resulting in $n=500$ samples. For Hypothesis~\ref{hyp:celebrity}, we include Variant 5, resulting in $n=100$ samples.

\subsection{Syllogistic Fallacy}

For Hypothesis~\ref{hyp:sets}, we randomly sample an entity \texttt{\{random\_object\}} from a curated list of objects from \citet{eslkidsvocabulary} and use the following few-shot prompt to generate $n=200$ problems:

\begin{quote}
\texttt{
Fill in the blanks in the following template. Do not output anything else.\\
All [objects] are [category].\\
Some [category]s [characteristic traits of this category].\\
Therefore some [same objects as before] [characteristic traits this category].\\
Make sure that the characteristic traits of this category only fit for a subset of this category but not for all.\\
For example:\\
All carrots are vegetables.\\
Some vegetables are rich in fiber.\\
Therefore, some carrots are rich in fiber.\\
All roses are flowers. \\
Some flowers fade quickly. \\
Therefore some roses fade quickly.\\
All actors are performers.\\
Some performers are skilled in improvisation.\\
Therefore some actors are skilled in improvisation.\\
All \{random\_object\} are 
}
\end{quote}

The common U.S. news media sources we used to perturb these problems in Hypothesis~\ref{hyp:framing} are taken from \cite{wikipedia2024newsmediausa, pew2011top25}. This also results in $n=200$ samples.

\newpage
\section{Additional Experiment Results}\label{app:full_results}

\subsection{Hypothesis 1}

\begin{figure*}[h]
    \centering
    \includegraphics[width=\textwidth]{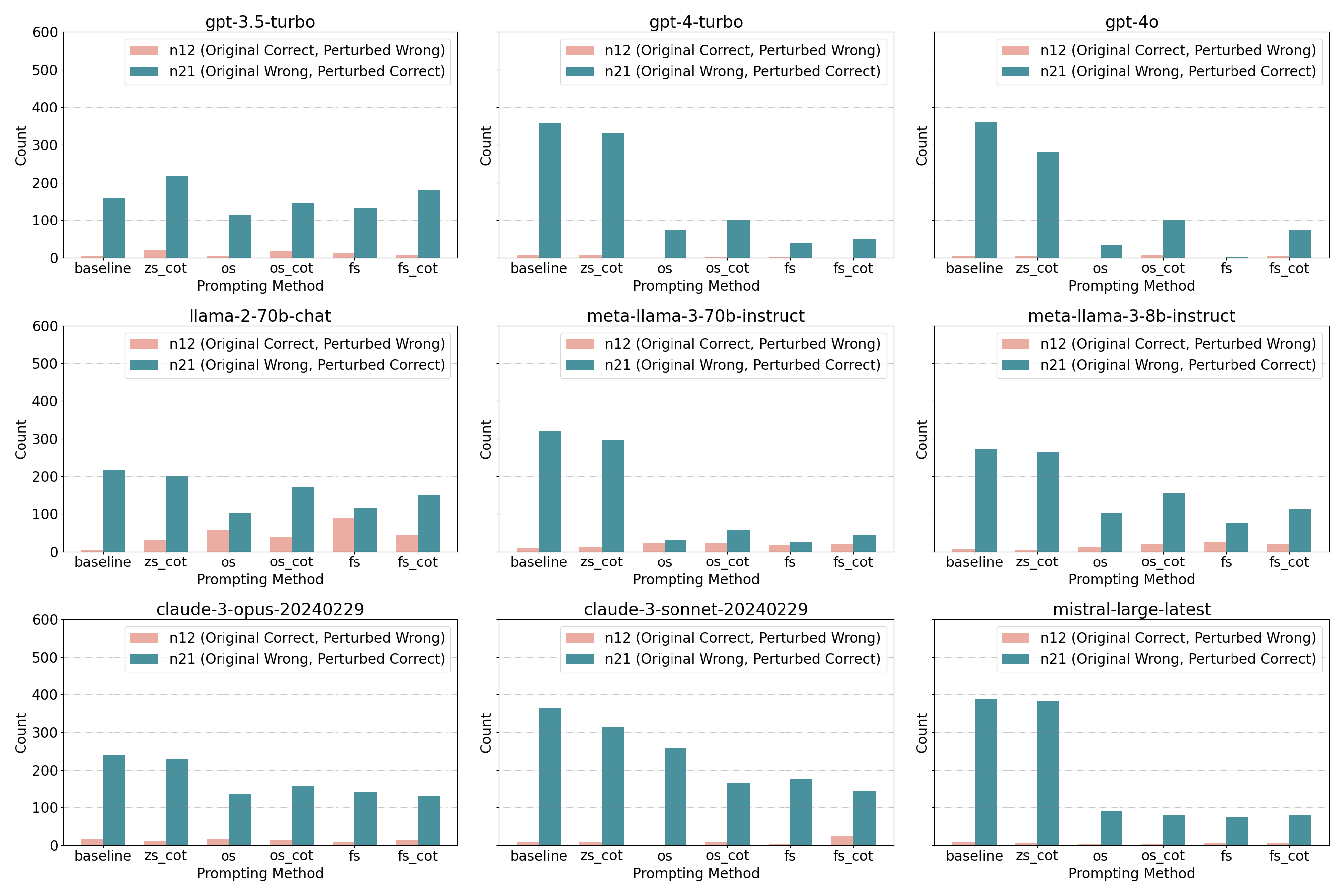}
    \caption{Full experimental results for Hypothesis \ref{hyp:lost_in_context} ($n = 400$). The perturbed problems alternate options contextually relevant to the problem statements to irrelevant ones. We run all different prompt methods. To reject the null, we expect $\textcolor{goldenGroupColor}{n12} < \textcolor{controlGroupColor}{n21}.$ We conclude that LLMs fail to reason against contextually misleading options in conjunction fallacy problems.}
    \label{fig:h1_full}
\end{figure*}

\begin{figure*}[h]
    \centering
    \includegraphics[width=\textwidth]{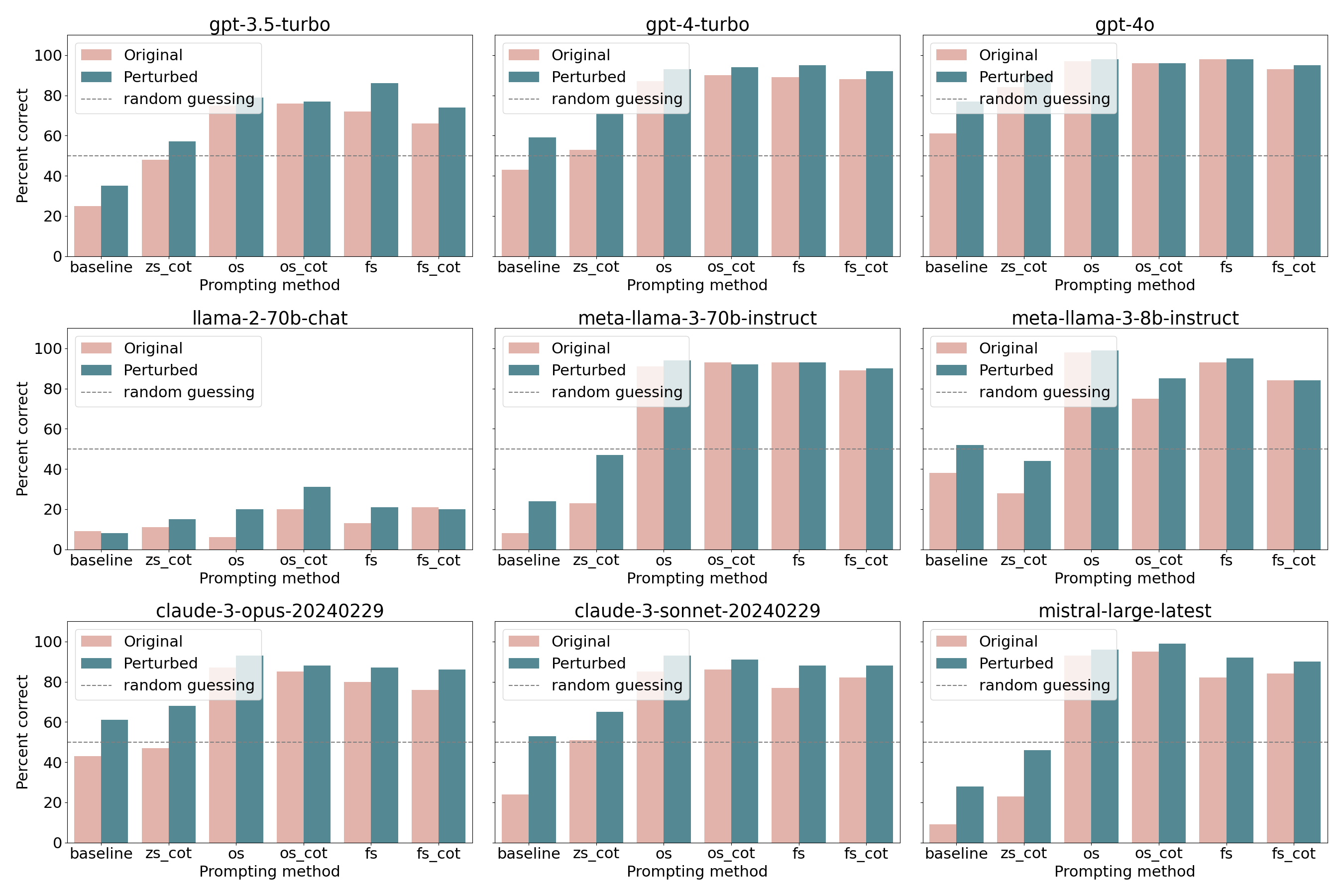}
    \caption{Comparison of the accuracy scores between the original and perturbed problems for Hypothesis \ref{hyp:lost_in_context}.}
    \label{fig:h1_full_acc}
\end{figure*}

The full experimental results for Hypothesis \ref{hyp:lost_in_context} are shown in Figure \ref{fig:h1_full}, \ref{fig:h1_full_acc} and Table \ref{table:h1}.

\begin{longtable}{llrrrrrr}
\caption{Full Experimental results for Hypothesis \ref{hyp:lost_in_context}. Note that in our experiments, $n^* = n_{21} + n_{12}$ is not equal to the number of data samples $n$. Here, $n_{12}$ denotes the instances where the LLM correctly answers the original problem but fails on the perturbed version, and $n_{21}$ denotes the opposite scenario. Thus, a large value of $n^*$ happens only when the LLM makes many mistakes. Specifically, GPT-4o in this table shows $n^* = 1$ with few-shots learning. We find that GPT-4o is excellent in answering these problems with few-show exemplars, achieving near-perfect accuracy of almost $100\%$, so their $n_{12}$ and $n_{21}$ values are pretty small. This high accuracy, however, only lead us to fail to reject this particular instantiation of the null hypothesis, but not in other situations. While we could increase the sample size from the current $n$ to potentialy observe more errors,  thus a higher $n^*$ in scenarios involving state-of-the-art LMs with few-shot learning, our rejection of the null hypothesis under other scenarios when tested against the GPT-4o is sufficient to argue that LLMs are not yet genuine reasoners.}\label{table:h1} \\
\toprule
model & prompting method & $n_{12}$ & $n_{21}$ & $n^*$ & z-stat & p-value & reject \\
\midrule
\endfirsthead

\multicolumn{8}{c}%
{\tablename\ \thetable\ -- \textit{Continued from previous page}} \\
\toprule
model & prompting method & $n_{12}$ & $n_{21}$ & $n^*$ & z-stat & p-value & reject \\
\midrule
\endhead

\midrule \multicolumn{8}{r}{\textit{Continued on next page}} \\
\endfoot

\bottomrule
\endlastfoot

gpt-3.5-turbo & baseline & 4 & 160 & 164 & 12.181553 & 0.000000 & True \\
gpt-3.5-turbo & zs-cot & 19 & 218 & 237 & 12.926439 & 0.000000 & True \\
gpt-3.5-turbo & os & 3 & 115 & 118 & 10.310436 & 0.000000 & True \\
gpt-3.5-turbo & os-cot & 17 & 147 & 164 & 10.151295 & 0.000000 & True \\
gpt-3.5-turbo & fs & 12 & 132 & 144 & 10.000000 & 0.000000 & True \\
gpt-3.5-turbo & fs-cot & 6 & 180 & 186 & 12.758299 & 0.000000 & True \\
gpt-4-turbo & baseline & 7 & 357 & 364 & 18.344985 & 0.000000 & True \\
gpt-4-turbo & zs-cot & 6 & 331 & 337 & 17.703878 & 0.000000 & True \\
gpt-4-turbo & os & 0 & 73 & 73 & 8.544004 & 0.000000 & True \\
gpt-4-turbo & os-cot & 1 & 101 & 102 & 9.901475 & 0.000000 & True \\
gpt-4-turbo & fs & 1 & 38 & 39 & 5.924742 & 0.000000 & True \\
gpt-4-turbo & fs-cot & 0 & 50 & 50 & 7.071068 & 0.000000 & True \\
gpt-4o & baseline & 5 & 360 & 365 & 18.581549 & 0.000000 & True \\
gpt-4o & zs-cot & 3 & 281 & 284 & 16.496265 & 0.000000 & True \\
gpt-4o & os & 0 & 33 & 33 & 5.744563 & 0.000000 & True \\
gpt-4o & os-cot & 8 & 101 & 109 & 8.907784 & 0.000000 & True \\
gpt-4o & fs & 0 & 1 & 1 & 1.000000 & 0.158655 & False \\
gpt-4o & fs-cot & 3 & 72 & 75 & 7.967434 & 0.000000 & True \\
llama-2-70b-chat & baseline & 3 & 215 & 218 & 14.358452 & 0.000000 & True \\
llama-2-70b-chat & zs-cot & 30 & 199 & 229 & 11.167834 & 0.000000 & True \\
llama-2-70b-chat & os & 56 & 101 & 157 & 3.591391 & 0.000181 & True \\
llama-2-70b-chat & os-cot & 38 & 170 & 208 & 9.152553 & 0.000000 & True \\
llama-2-70b-chat & fs & 90 & 115 & 205 & 1.746076 & 0.042775 & True \\
llama-2-70b-chat & fs-cot & 43 & 150 & 193 & 7.702029 & 0.000000 & True \\
meta-llama-3-70b-instruct & baseline & 10 & 321 & 331 & 17.094106 & 0.000000 & True \\
meta-llama-3-70b-instruct & zs-cot & 12 & 296 & 308 & 16.182402 & 0.000000 & True \\
meta-llama-3-70b-instruct & os & 22 & 32 & 54 & 1.360828 & 0.090122 & False \\
meta-llama-3-70b-instruct & os-cot & 22 & 58 & 80 & 4.024922 & 0.000032 & True \\
meta-llama-3-70b-instruct & fs & 18 & 26 & 44 & 1.206045 & 0.116049 & False \\
meta-llama-3-70b-instruct & fs-cot & 19 & 44 & 63 & 3.149704 & 0.000883 & True \\
meta-llama-3-8b-instruct & baseline & 8 & 272 & 280 & 15.777018 & 0.000000 & True \\
meta-llama-3-8b-instruct & zs-cot & 5 & 263 & 268 & 15.759858 & 0.000000 & True \\
meta-llama-3-8b-instruct & os & 12 & 102 & 114 & 8.429272 & 0.000000 & True \\
meta-llama-3-8b-instruct & os-cot & 19 & 154 & 173 & 10.263860 & 0.000000 & True \\
meta-llama-3-8b-instruct & fs & 26 & 77 & 103 & 5.025179 & 0.000000 & True \\
meta-llama-3-8b-instruct & fs-cot & 20 & 112 & 132 & 8.007572 & 0.000000 & True \\
claude-3-opus-20240229 & baseline & 17 & 241 & 258 & 13.945631 & 0.000000 & True \\
claude-3-opus-20240229 & zs-cot & 10 & 229 & 239 & 14.165932 & 0.000000 & True \\
claude-3-opus-20240229 & os & 15 & 136 & 151 & 9.846840 & 0.000000 & True \\
claude-3-opus-20240229 & os-cot & 13 & 157 & 170 & 11.044296 & 0.000000 & True \\
claude-3-opus-20240229 & fs & 9 & 140 & 149 & 10.731938 & 0.000000 & True \\
claude-3-opus-20240229 & fs-cot & 14 & 129 & 143 & 9.616783 & 0.000000 & True \\
claude-3-sonnet-20240229 & baseline & 8 & 364 & 372 & 18.457740 & 0.000000 & True \\
claude-3-sonnet-20240229 & zs-cot & 8 & 313 & 321 & 17.023440 & 0.000000 & True \\
claude-3-sonnet-20240229 & os & 0 & 258 & 258 & 16.062378 & 0.000000 & True \\
claude-3-sonnet-20240229 & os-cot & 9 & 165 & 174 & 11.826329 & 0.000000 & True \\
claude-3-sonnet-20240229 & fs & 3 & 175 & 178 & 12.891945 & 0.000000 & True \\
claude-3-sonnet-20240229 & fs-cot & 24 & 143 & 167 & 9.208496 & 0.000000 & True \\
mistral-large-latest & baseline & 8 & 388 & 396 & 19.095718 & 0.000000 & True \\
mistral-large-latest & zs-cot & 5 & 384 & 389 & 19.216063 & 0.000000 & True \\
mistral-large-latest & os & 3 & 91 & 94 & 9.076507 & 0.000000 & True \\
mistral-large-latest & os-cot & 4 & 79 & 83 & 8.232319 & 0.000000 & True \\
mistral-large-latest & fs & 5 & 74 & 79 & 7.763107 & 0.000000 & True \\
mistral-large-latest & fs-cot & 5 & 79 & 84 & 8.074062 & 0.000000 & True \\
\end{longtable}

\FloatBarrier

\FloatBarrier

\newpage
\subsection{Hypothesis 2}

\begin{figure*}[h]
    \centering
    \includegraphics[width=\textwidth]{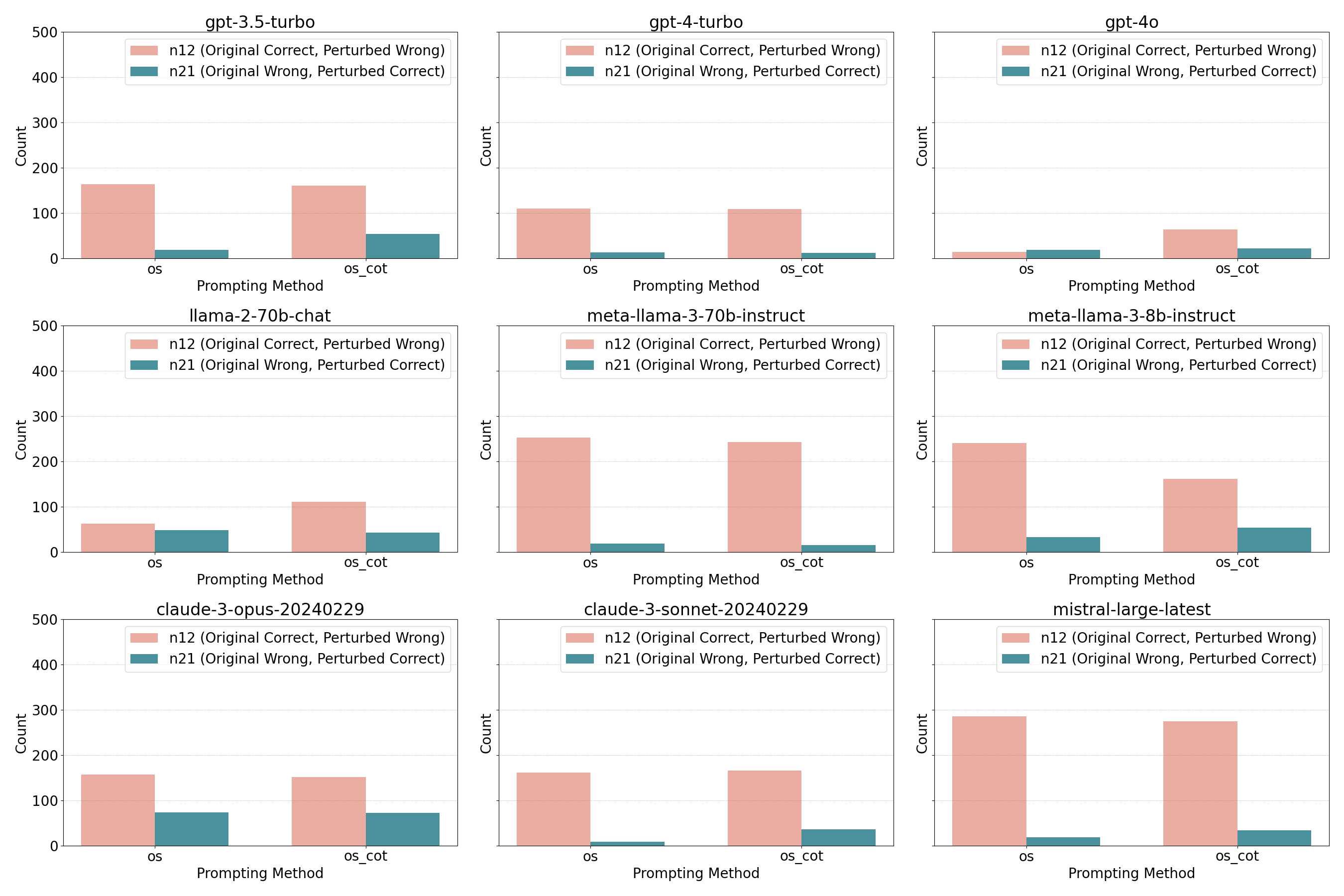}
    \caption{Full experimental results for Hypothesis \ref{hyp:linda_bob} ($n = 500$). The perturbed problems alternate the name classic "Linda" to "Bob" in in-context learning exemplars. We run one-shot with and without chain-of-thought prompts. To reject the null, we expect $\textcolor{goldenGroupColor}{n12} > \textcolor{controlGroupColor}{n21}$. We conclude that LLMs possess a strong token bias toward the name "Linda", which frequently appears in classic literature.}
    \label{fig:h2_full}
\end{figure*}

\begin{figure*}[h]
    \centering
    \includegraphics[width=\textwidth]{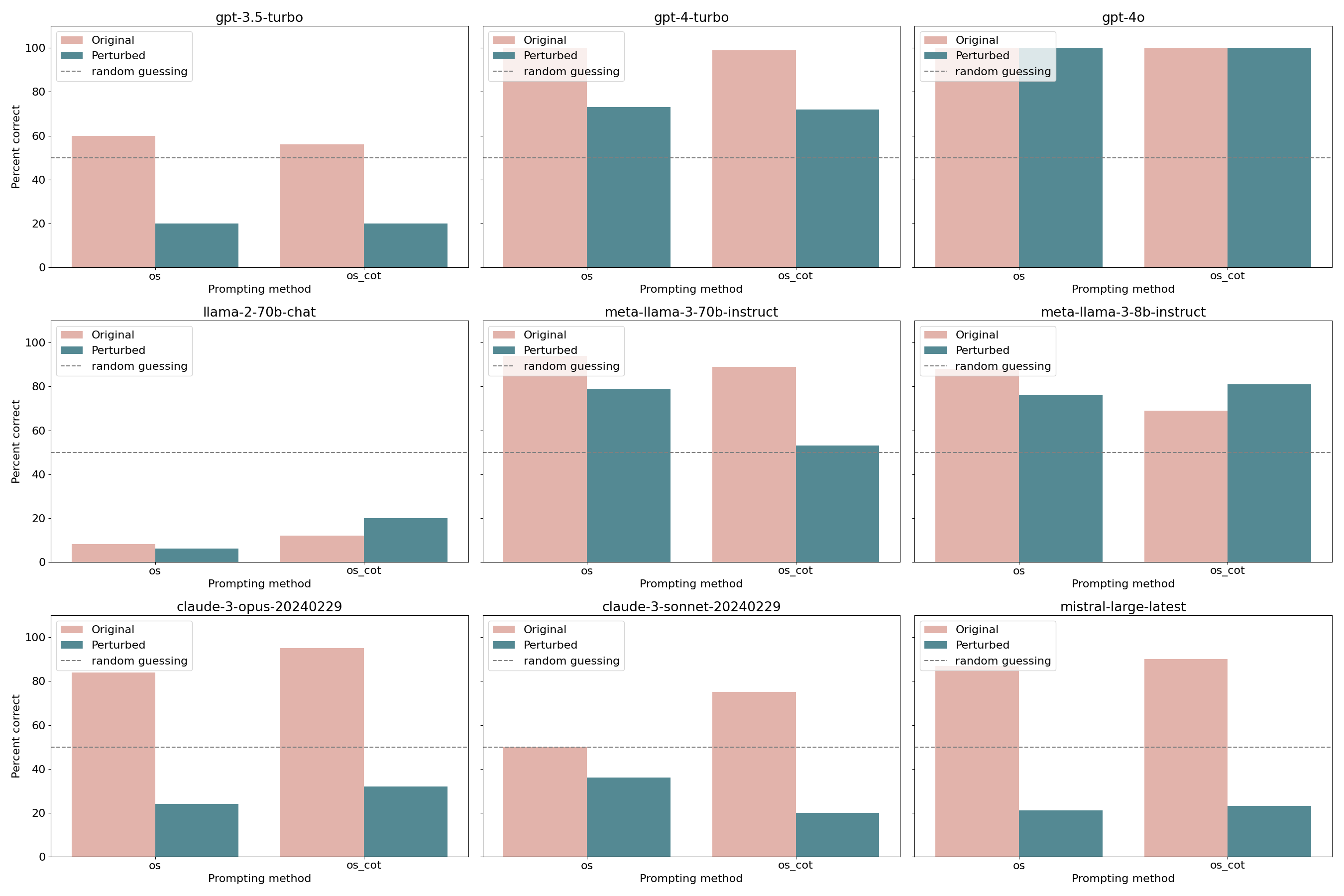}
    \caption{Comparison of the accuracy scores between the original and perturbed problems for Hypothesis \ref{hyp:linda_bob}.}
    \label{fig:h2_full_acc}
\end{figure*}

The full experimental results for Hypothesis \ref{hyp:linda_bob} are shown in Figure \ref{fig:h2_full}, \ref{fig:h2_full_acc} and Table \ref{table:h2}.

\begin{longtable}{llrrrrrr}
\caption{Full experimental results for Hypothesis \ref{hyp:linda_bob}}
\label{table:h2} \\
\toprule
model & prompting method & $n_{12}$ & $n_{21}$ & $n^*$ & z-stat & p-value & reject \\
\midrule
\endfirsthead

\multicolumn{8}{c}%
{\tablename\ \thetable\ -- \textit{Continued from previous page}} \\
\toprule
model & prompting method & $n_{12}$ & $n_{21}$ & $n^*$ & z-stat & p-value & reject \\
\midrule
\endhead

\midrule \multicolumn{8}{r}{\textit{Continued on next page}} \\
\endfoot

\bottomrule
\endlastfoot

gpt-3.5-turbo & os & 164 & 18 & 182 & -10.822240 & 0.000000 & True \\
gpt-3.5-turbo & os-cot & 160 & 54 & 214 & -7.246011 & 0.000000 & True \\
gpt-4-turbo & os & 110 & 13 & 123 & -8.746195 & 0.000000 & True \\
gpt-4-turbo & os-cot & 109 & 12 & 121 & -8.818182 & 0.000000 & True \\
gpt-4o & os & 14 & 18 & 32 & 0.707107 & 0.760250 & False \\
gpt-4o & os-cot & 64 & 22 & 86 & -4.528976 & 0.000003 & True \\
llama-2-70b-chat & os & 62 & 48 & 110 & -1.334848 & 0.096314 & False \\
llama-2-70b-chat & os-cot & 111 & 43 & 154 & -5.479596 & 0.000000 & True \\
meta-llama-3-70b-instruct & os & 253 & 18 & 271 & -14.275233 & 0.000000 & True \\
meta-llama-3-70b-instruct & os-cot & 243 & 15 & 258 & -14.194660 & 0.000000 & True \\
meta-llama-3-8b-instruct & os & 241 & 33 & 274 & -12.565740 & 0.000000 & True \\
meta-llama-3-8b-instruct & os-cot & 162 & 54 & 216 & -7.348469 & 0.000000 & True \\
claude-3-opus-20240229 & os & 157 & 73 & 230 & -5.538796 & 0.000000 & True \\
claude-3-opus-20240229 & os-cot & 151 & 72 & 223 & -5.290231 & 0.000000 & True \\
claude-3-sonnet-20240229 & os & 162 & 9 & 171 & -11.700202 & 0.000000 & True \\
claude-3-sonnet-20240229 & os-cot & 166 & 36 & 202 & -9.146768 & 0.000000 & True \\
mistral-large-latest & os & 286 & 18 & 304 & -15.370854 & 0.000000 & True \\
mistral-large-latest & os-cot & 275 & 34 & 309 & -13.710011 & 0.000000 & True \\
\end{longtable}

\FloatBarrier

\FloatBarrier

\newpage
\subsection{Hypothesis 3}

\begin{figure*}[h]
    \centering
    \includegraphics[width=\textwidth]{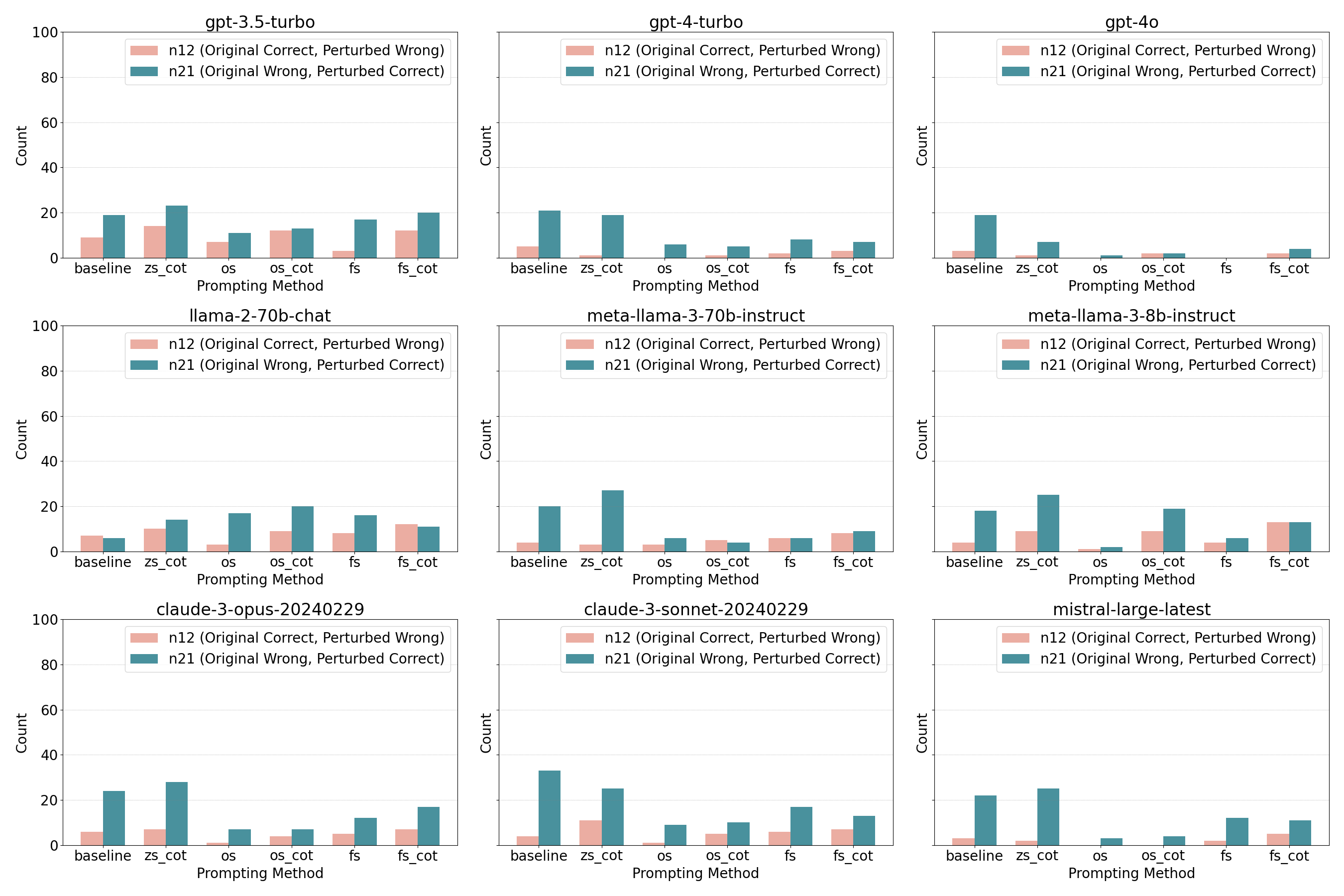}
    \caption{Full experimental results for Hypothesis \ref{hyp:celebrity} ($n = 100$). The perturbed problems alternate the celebrity name to a generic one in problem statements. We run all different prompt methods. To reject the null, we expect $\textcolor{goldenGroupColor}{n12} < \textcolor{controlGroupColor}{n21}.$ We conclude that LLMs are frequently misled by irrelevant celebrity names in problem statements that are irrelevant to logical essence.}
    \label{fig:h3_full}
\end{figure*}

\begin{figure*}[h]
    \centering
    \includegraphics[width=\textwidth]{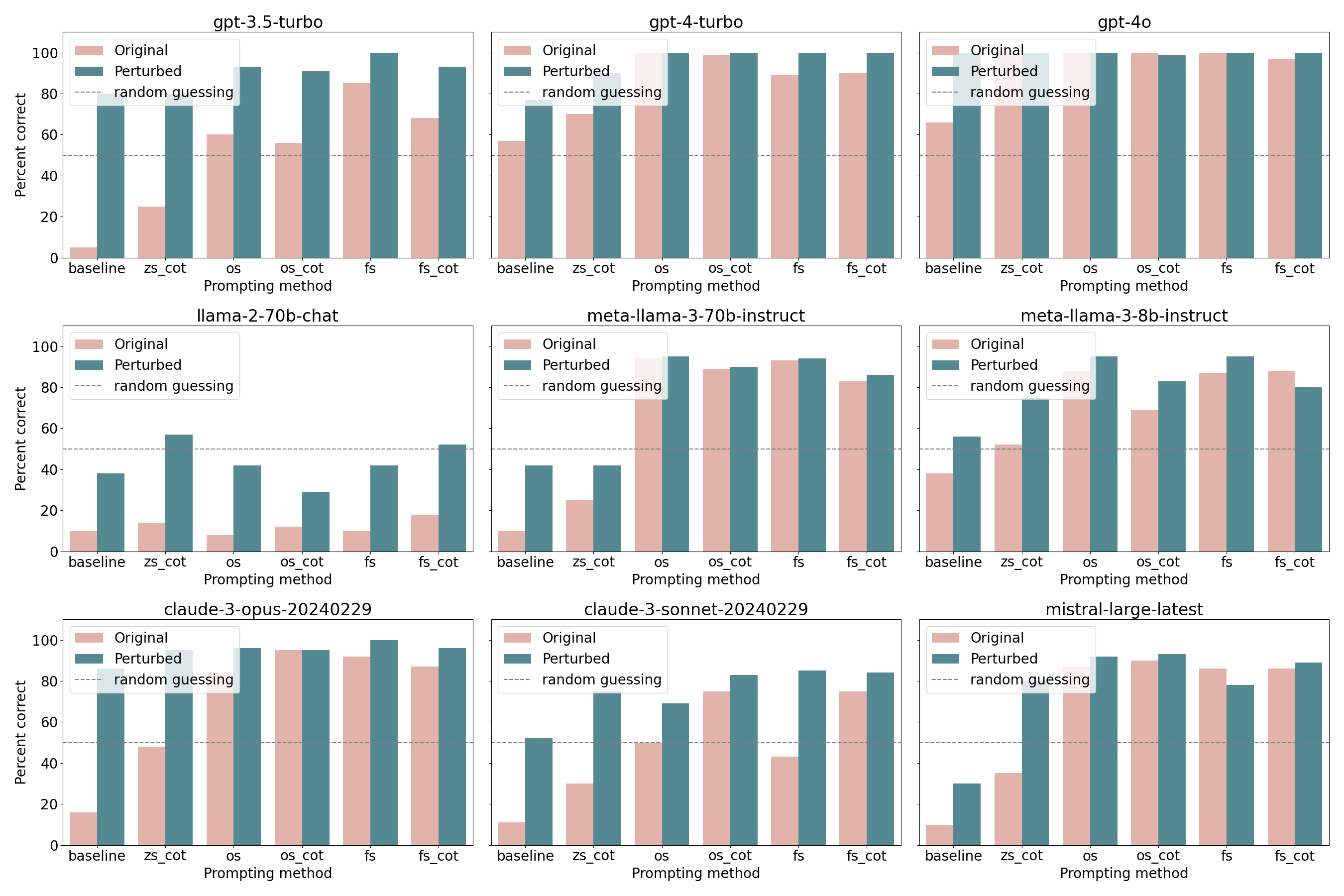}
    \caption{Comparison of the accuracy scores between the original and perturbed problems for Hypothesis \ref{hyp:celebrity}.}
    \label{fig:h3_full_acc}
\end{figure*}

The full experimental results for Hypothesis \ref{hyp:celebrity} are shown in Figure \ref{fig:h3_full}, \ref{fig:h3_full_acc} and Table \ref{table:h3}.

\begin{longtable}{llrrrrrr}
\caption{Full experimental results for Hypothesis \ref{hyp:celebrity}}
\label{table:h3} \\
\toprule
model & prompting method & $n_{12}$ & $n_{21}$ & $n^*$ & z-stat & p-value & reject \\
\midrule
\endfirsthead

\multicolumn{8}{c}%
{\tablename\ \thetable\ -- \textit{Continued from previous page}} \\
\toprule
model & prompting method & $n_{12}$ & $n_{21}$ & $n^*$ & z-stat & p-value & reject \\
\midrule
\endhead

\midrule \multicolumn{8}{r}{\textit{Continued on next page}} \\
\endfoot

\bottomrule
\endlastfoot

gpt-3.5-turbo & baseline & 9 & 19 & 28 & 1.889822 & 0.072141 & False \\
gpt-3.5-turbo & zs-cot & 14 & 23 & 37 & 1.479591 & 0.133569 & False \\
gpt-3.5-turbo & os & 7 & 11 & 18 & 0.942809 & 0.341537 & False \\
gpt-3.5-turbo & os-cot & 12 & 13 & 25 & 0.200000 & 0.516363 & False \\
gpt-3.5-turbo & fs & 3 & 17 & 20 & 3.130495 & 0.005798 & True \\
gpt-3.5-turbo & fs-cot & 12 & 20 & 32 & 1.414214 & 0.137003 & False \\
gpt-4-turbo & baseline & 5 & 21 & 26 & 3.137858 & 0.004595 & True \\
gpt-4-turbo & zs-cot & 1 & 19 & 20 & 4.024922 & 0.000270 & True \\
gpt-4-turbo & os & 0 & 6 & 6 & 2.449490 & 0.046875 & True \\
gpt-4-turbo & os-cot & 1 & 5 & 6 & 1.632993 & 0.178977 & False \\
gpt-4-turbo & fs & 2 & 8 & 10 & 1.897367 & 0.113582 & False \\
gpt-4-turbo & fs-cot & 3 & 7 & 10 & 1.264911 & 0.250845 & False \\
gpt-4o & baseline & 3 & 19 & 22 & 3.411211 & 0.003300 & True \\
gpt-4o & zs-cot & 1 & 7 & 8 & 2.121320 & 0.075938 & False \\
gpt-4o & os & 0 & 1 & 1 & 1.000000 & 0.562500 & False \\
gpt-4o & os-cot & 2 & 2 & 4 & 0.000000 & 0.727941 & False \\
gpt-4o & fs & 0 & 0 & 0 & 0.000000 & 1.000000 & False \\
gpt-4o & fs-cot & 2 & 4 & 6 & 0.816497 & 0.441964 & False \\
llama-2-70b-chat & baseline & 7 & 6 & 13 & -0.277350 & 0.736760 & False \\
llama-2-70b-chat & zs-cot & 10 & 14 & 24 & 0.816497 & 0.361423 & False \\
llama-2-70b-chat & os & 3 & 17 & 20 & 3.130495 & 0.005798 & True \\
llama-2-70b-chat & os-cot & 9 & 20 & 29 & 2.042649 & 0.055468 & False \\
llama-2-70b-chat & fs & 8 & 16 & 24 & 1.632993 & 0.136431 & False \\
llama-2-70b-chat & fs-cot & 12 & 11 & 23 & -0.208514 & 0.714075 & False \\
meta-llama-3-70b-instruct & baseline & 4 & 20 & 24 & 3.265986 & 0.004595 & True \\
meta-llama-3-70b-instruct & zs-cot & 3 & 27 & 30 & 4.381780 & 0.000106 & True \\
meta-llama-3-70b-instruct & os & 3 & 6 & 9 & 1.000000 & 0.351562 & False \\
meta-llama-3-70b-instruct & os-cot & 5 & 4 & 9 & -0.333333 & 0.760171 & False \\
meta-llama-3-70b-instruct & fs & 6 & 6 & 12 & 0.000000 & 0.675323 & False \\
meta-llama-3-70b-instruct & fs-cot & 8 & 9 & 17 & 0.242536 & 0.562500 & False \\
meta-llama-3-8b-instruct & baseline & 4 & 18 & 22 & 2.984810 & 0.009021 & True \\
meta-llama-3-8b-instruct & zs-cot & 9 & 25 & 34 & 2.743977 & 0.011706 & True \\
meta-llama-3-8b-instruct & os & 1 & 2 & 3 & 0.577350 & 0.562500 & False \\
meta-llama-3-8b-instruct & os-cot & 9 & 19 & 28 & 1.889822 & 0.072141 & False \\
meta-llama-3-8b-instruct & fs & 4 & 6 & 10 & 0.632456 & 0.473383 & False \\
meta-llama-3-8b-instruct & fs-cot & 13 & 13 & 26 & 0.000000 & 0.562500 & False \\
claude-3-opus-20240229 & baseline & 6 & 24 & 30 & 3.286335 & 0.003426 & True \\
claude-3-opus-20240229 & zs-cot & 7 & 28 & 35 & 3.549648 & 0.001736 & True \\
claude-3-opus-20240229 & os & 1 & 7 & 8 & 2.121320 & 0.075938 & False \\
claude-3-opus-20240229 & os-cot & 4 & 7 & 11 & 0.904534 & 0.361423 & False \\
claude-3-opus-20240229 & fs & 5 & 12 & 17 & 1.697749 & 0.133569 & False \\
claude-3-opus-20240229 & fs-cot & 7 & 17 & 24 & 2.041241 & 0.075030 & False \\
claude-3-sonnet-20240229 & baseline & 4 & 33 & 37 & 4.767571 & 0.000050 & True \\
claude-3-sonnet-20240229 & zs-cot & 11 & 25 & 36 & 2.333333 & 0.033127 & True \\
claude-3-sonnet-20240229 & os & 1 & 9 & 10 & 2.529822 & 0.034122 & True \\
claude-3-sonnet-20240229 & os-cot & 5 & 10 & 15 & 1.290994 & 0.226318 & False \\
claude-3-sonnet-20240229 & fs & 6 & 17 & 23 & 2.293659 & 0.049296 & True \\
claude-3-sonnet-20240229 & fs-cot & 7 & 13 & 20 & 1.341641 & 0.203021 & False \\
mistral-large-latest & baseline & 3 & 22 & 25 & 3.800000 & 0.000781 & True \\
mistral-large-latest & zs-cot & 2 & 25 & 27 & 4.426352 & 0.000106 & True \\
mistral-large-latest & os & 0 & 3 & 3 & 1.732051 & 0.198529 & False \\
mistral-large-latest & os-cot & 0 & 4 & 4 & 2.000000 & 0.125000 & False \\
mistral-large-latest & fs & 2 & 12 & 14 & 2.672612 & 0.023291 & True \\
mistral-large-latest & fs-cot & 5 & 11 & 16 & 1.500000 & 0.177283 & False \\
\end{longtable}

\FloatBarrier

\FloatBarrier

\newpage
\subsection{Hypothesis 4}

\begin{figure*}[h]
    \centering
    \includegraphics[width=\textwidth]{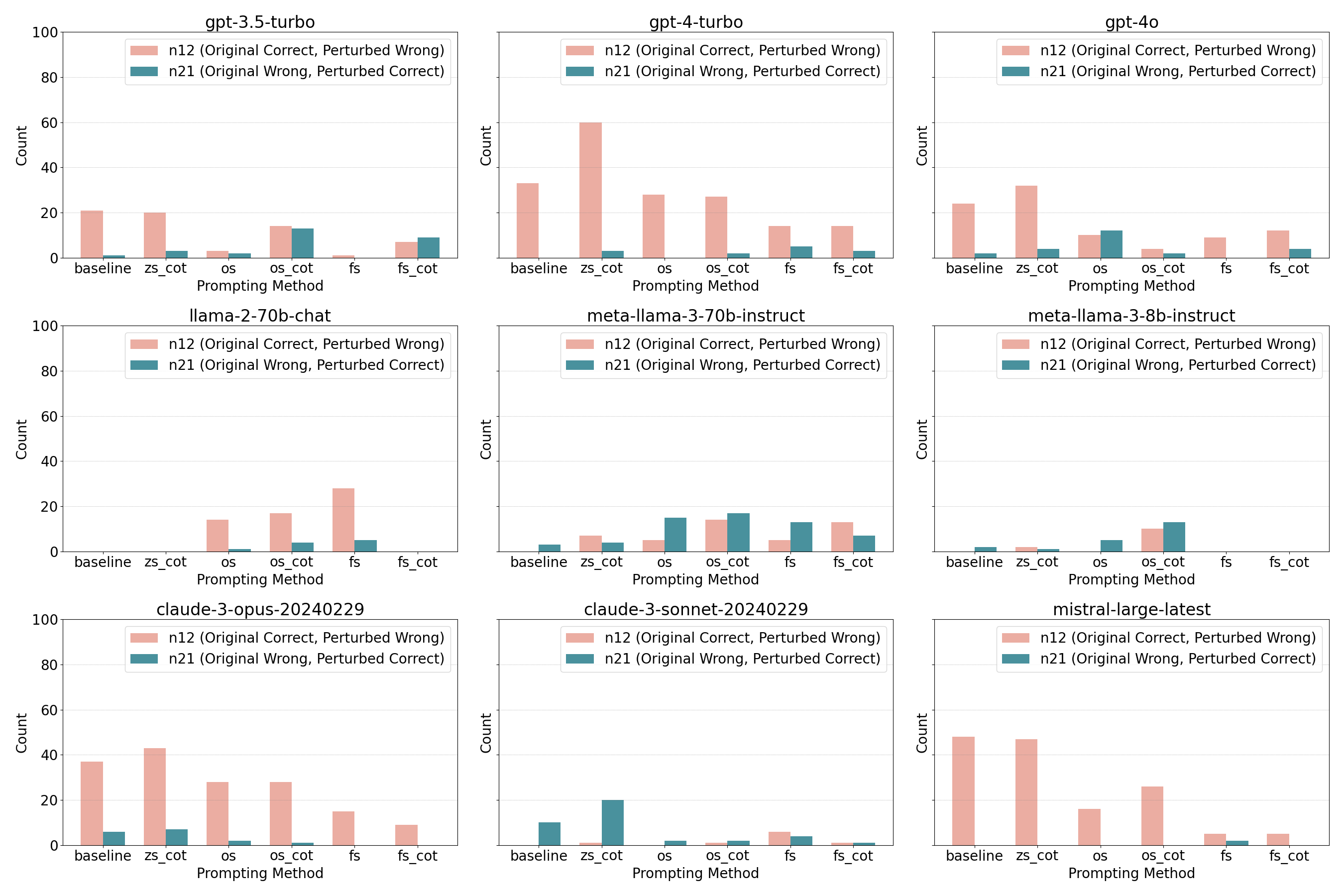}
    \caption{Full experimental results for Hypothesis \ref{hyp:sets} ($n = 200$). The perturbed problems alternate tokens "All" and "Some" to different but equivalent expressions in syllogisms. We run all different prompt methods. To reject the null, we expect $\textcolor{goldenGroupColor}{n12} > \textcolor{controlGroupColor}{n21}.$ We conclude that most LLMs rely on patterns \textit{"All..., Some..., Some..."} for reasoning about syllogism.}
    \label{fig:h4_full}
\end{figure*}

\begin{figure*}[h]
    \centering
    \includegraphics[width=\textwidth]{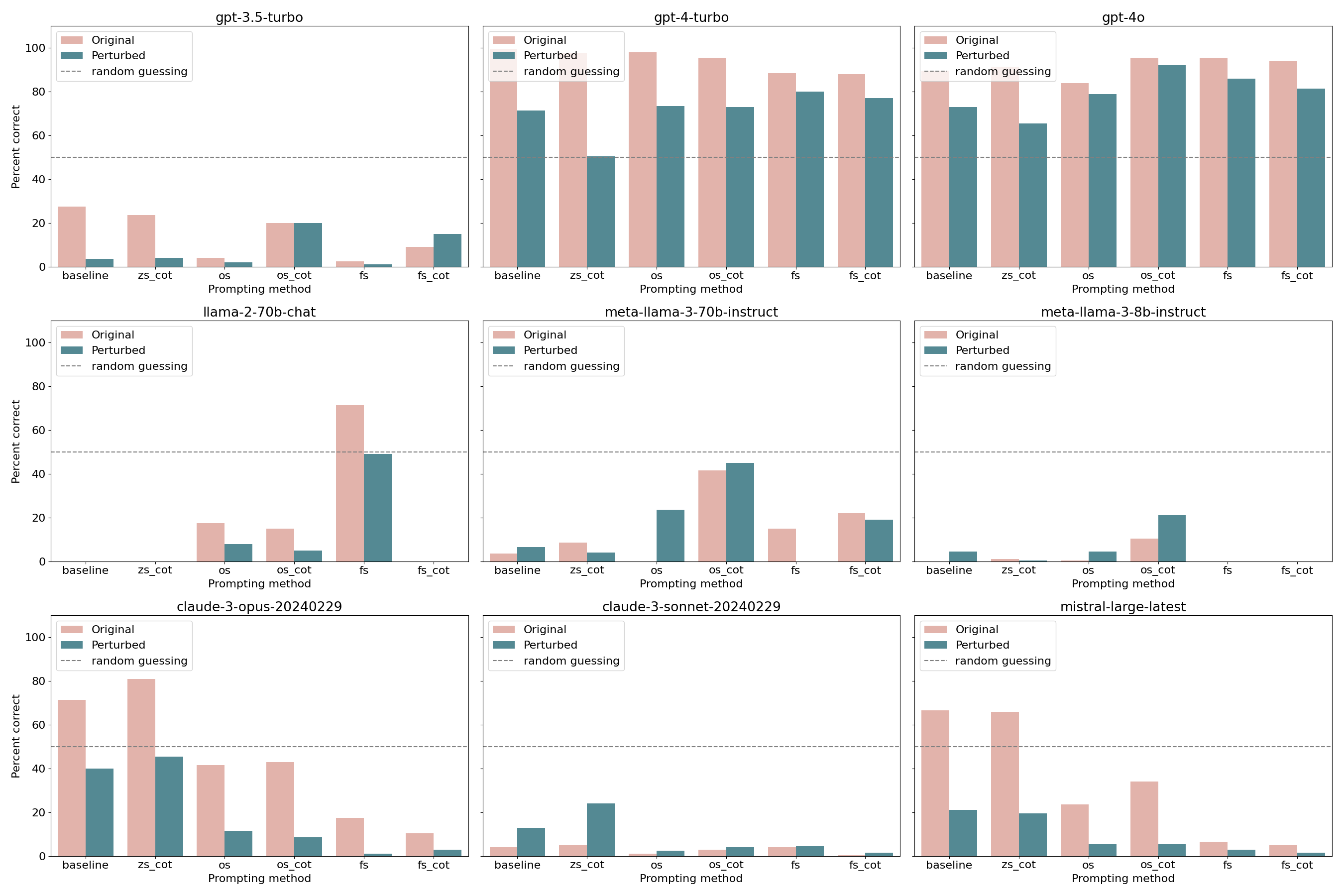}
    \caption{Comparison of the accuracy scores between the original and perturbed problems for Hypothesis \ref{hyp:sets}.}
    \label{fig:h4_full_acc}
\end{figure*}

The full experimental results for Hypothesis \ref{hyp:sets} are shown in Figure \ref{fig:h4_full}, \ref{fig:h4_full_acc} and Table \ref{table:h4}.

\begin{longtable}{llrrrrrr}
\caption{Full experimental results for Hypothesis \ref{hyp:sets}}
\label{table:h4} \\
\toprule
model & prompting method & $n_{12}$ & $n_{21}$ & $n^*$ & z-stat & p-value & reject \\
\midrule
\endfirsthead

\multicolumn{8}{c}%
{\tablename\ \thetable\ -- \textit{Continued from previous page}} \\
\toprule
model & prompting method & $n_{12}$ & $n_{21}$ & $n^*$ & z-stat & p-value & reject \\
\midrule
\endhead

\midrule \multicolumn{8}{r}{\textit{Continued on next page}} \\
\endfoot

\bottomrule
\endlastfoot

gpt-3.5-turbo & baseline & 21 & 1 & 22 & -4.264014 & 0.000023 & True \\
gpt-3.5-turbo & zs-cot & 20 & 3 & 23 & -3.544745 & 0.000732 & True \\
gpt-3.5-turbo & os & 3 & 2 & 5 & -0.447214 & 0.771429 & False \\
gpt-3.5-turbo & os-cot & 14 & 13 & 27 & -0.192450 & 0.714985 & False \\
gpt-3.5-turbo & fs & 1 & 0 & 1 & -1.000000 & 0.771429 & False \\
gpt-3.5-turbo & fs-cot & 7 & 9 & 16 & 0.500000 & 1.000000 & False \\
gpt-4-turbo & baseline & 33 & 0 & 33 & -5.744563 & 0.000000 & True \\
gpt-4-turbo & zs-cot & 60 & 3 & 63 & -7.181325 & 0.000000 & True \\
gpt-4-turbo & os & 28 & 0 & 28 & -5.291503 & 0.000001 & True \\
gpt-4-turbo & os-cot & 27 & 2 & 29 & -4.642383 & 0.000008 & True \\
gpt-4-turbo & fs & 14 & 5 & 19 & -2.064742 & 0.068654 & False \\
gpt-4-turbo & fs-cot & 14 & 3 & 17 & -2.667892 & 0.014939 & True \\
gpt-4o & baseline & 24 & 2 & 26 & -4.314555 & 0.000031 & True \\
gpt-4o & zs-cot & 32 & 4 & 36 & -4.666667 & 0.000008 & True \\
gpt-4o & os & 10 & 12 & 22 & 0.426401 & 1.000000 & False \\
gpt-4o & os-cot & 4 & 2 & 6 & -0.816497 & 0.618750 & False \\
gpt-4o & fs & 9 & 0 & 9 & -3.000000 & 0.005022 & True \\
gpt-4o & fs-cot & 12 & 4 & 16 & -2.000000 & 0.079767 & False \\
llama-2-70b-chat & baseline & 0 & 0 & 0 & 0.000000 & 1.000000 & False \\
llama-2-70b-chat & zs-cot & 0 & 0 & 0 & 0.000000 & 1.000000 & False \\
llama-2-70b-chat & os & 14 & 1 & 15 & -3.356586 & 0.001388 & True \\
llama-2-70b-chat & os-cot & 17 & 4 & 21 & -2.836833 & 0.008833 & True \\
llama-2-70b-chat & fs & 28 & 5 & 33 & -4.003786 & 0.000099 & True \\
llama-2-70b-chat & fs-cot & 0 & 0 & 0 & 0.000000 & 1.000000 & False \\
meta-llama-3-70b-instruct & baseline & 0 & 3 & 3 & 1.732051 & 1.000000 & False \\
meta-llama-3-70b-instruct & zs-cot & 7 & 4 & 11 & -0.904534 & 0.510978 & False \\
meta-llama-3-70b-instruct & os & 5 & 15 & 20 & 2.236068 & 1.000000 & False \\
meta-llama-3-70b-instruct & os-cot & 14 & 17 & 31 & 0.538816 & 1.000000 & False \\
meta-llama-3-70b-instruct & fs & 5 & 13 & 18 & 1.885618 & 1.000000 & False \\
meta-llama-3-70b-instruct & fs-cot & 13 & 7 & 20 & -1.341641 & 0.263176 & False \\
meta-llama-3-8b-instruct & baseline & 0 & 2 & 2 & 1.414214 & 1.000000 & False \\
meta-llama-3-8b-instruct & zs-cot & 2 & 1 & 3 & -0.577350 & 0.771429 & False \\
meta-llama-3-8b-instruct & os & 0 & 5 & 5 & 2.236068 & 1.000000 & False \\
meta-llama-3-8b-instruct & os-cot & 10 & 13 & 23 & 0.625543 & 1.000000 & False \\
meta-llama-3-8b-instruct & fs & 0 & 0 & 0 & 0.000000 & 1.000000 & False \\
meta-llama-3-8b-instruct & fs-cot & 0 & 0 & 0 & 0.000000 & 1.000000 & False \\
claude-3-opus-20240229 & baseline & 37 & 6 & 43 & -4.727456 & 0.000006 & True \\
claude-3-opus-20240229 & zs-cot & 43 & 7 & 50 & -5.091169 & 0.000001 & True \\
claude-3-opus-20240229 & os & 28 & 2 & 30 & -4.746929 & 0.000006 & True \\
claude-3-opus-20240229 & os-cot & 28 & 1 & 29 & -5.013774 & 0.000002 & True \\
claude-3-opus-20240229 & fs & 15 & 0 & 15 & -3.872983 & 0.000099 & True \\
claude-3-opus-20240229 & fs-cot & 9 & 0 & 9 & -3.000000 & 0.005022 & True \\
claude-3-sonnet-20240229 & baseline & 0 & 10 & 10 & 3.162278 & 1.000000 & False \\
claude-3-sonnet-20240229 & zs-cot & 1 & 20 & 21 & 4.146140 & 1.000000 & False \\
claude-3-sonnet-20240229 & os & 0 & 2 & 2 & 1.414214 & 1.000000 & False \\
claude-3-sonnet-20240229 & os-cot & 1 & 2 & 3 & 0.577350 & 1.000000 & False \\
claude-3-sonnet-20240229 & fs & 6 & 4 & 10 & -0.632456 & 0.656628 & False \\
claude-3-sonnet-20240229 & fs-cot & 1 & 1 & 2 & 0.000000 & 1.000000 & False \\
mistral-large-latest & baseline & 48 & 0 & 48 & -6.928203 & 0.000000 & True \\
mistral-large-latest & zs-cot & 47 & 0 & 47 & -6.855655 & 0.000000 & True \\
mistral-large-latest & os & 16 & 0 & 16 & -4.000000 & 0.000055 & True \\
mistral-large-latest & os-cot & 26 & 0 & 26 & -5.099020 & 0.000001 & True \\
mistral-large-latest & fs & 5 & 2 & 7 & -1.133893 & 0.436942 & False \\
mistral-large-latest & fs-cot & 5 & 0 & 5 & -2.236068 & 0.068654 & False \\
\end{longtable}

\FloatBarrier

\FloatBarrier

\newpage
\subsection{Hypothesis 5}

\begin{figure*}[h]
    \centering
    \includegraphics[width=\textwidth]{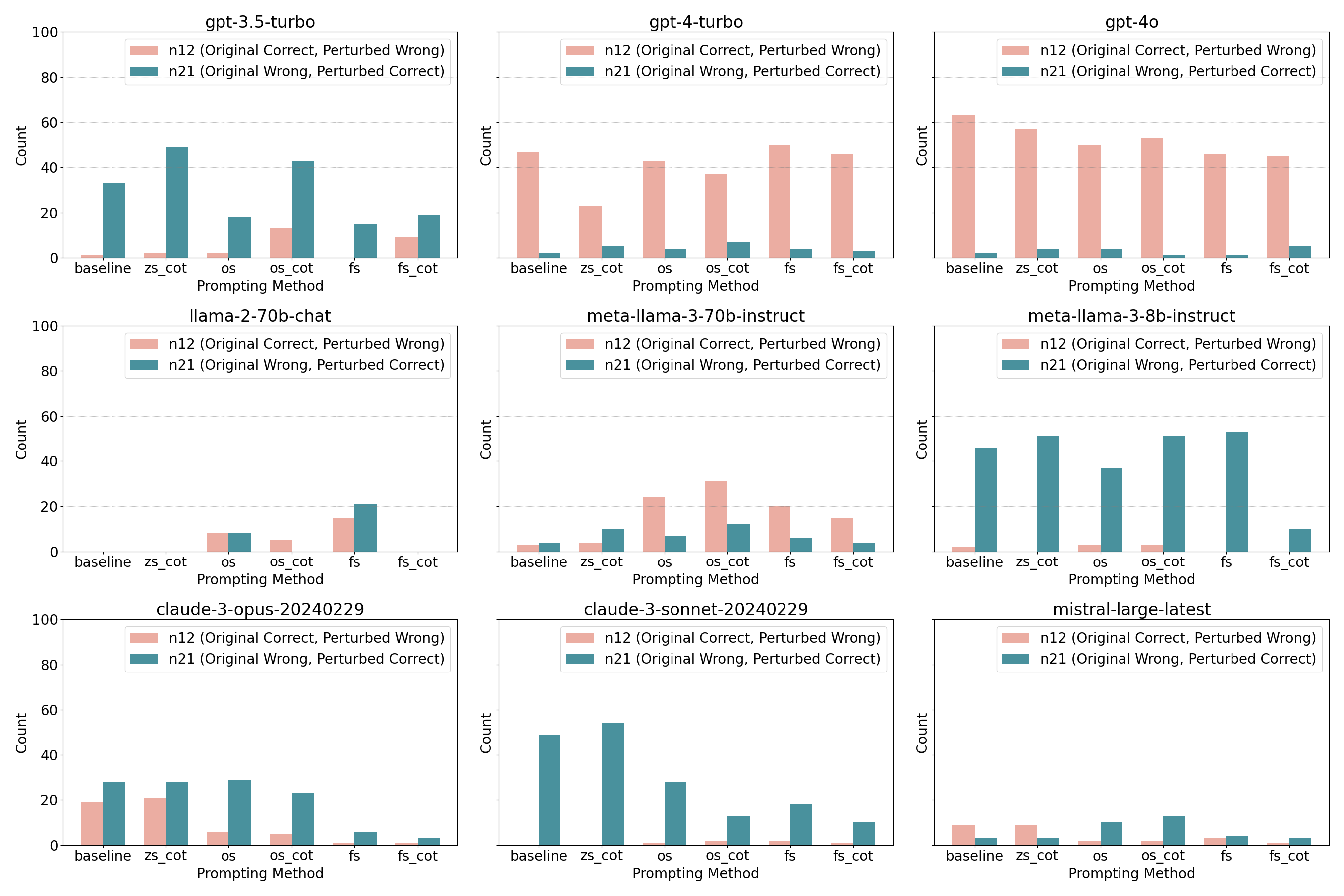}
    \caption{Full experimental results for Hypothesis \ref{hyp:framing}, sets-original vs. sets-framing ($n = 200$). The perturbed problems add the names of trustworthy news agencies and universities to alter the narratives of syllogisms. Both the original and perturbed problems have classic patterns \textit{"All..., Some..., Some..."} already removed to ensure a single confounder for token bias analysis. We run all different prompt methods. To reject the null, we expect $\textcolor{goldenGroupColor}{n12} > \textcolor{controlGroupColor}{n21}.$ We conclude that LLMs might be misled by reputable names irrelevant to the logical structure.}
    \label{fig:h5_full}
\end{figure*}

\begin{figure*}[h]
    \centering
    \includegraphics[width=\textwidth]{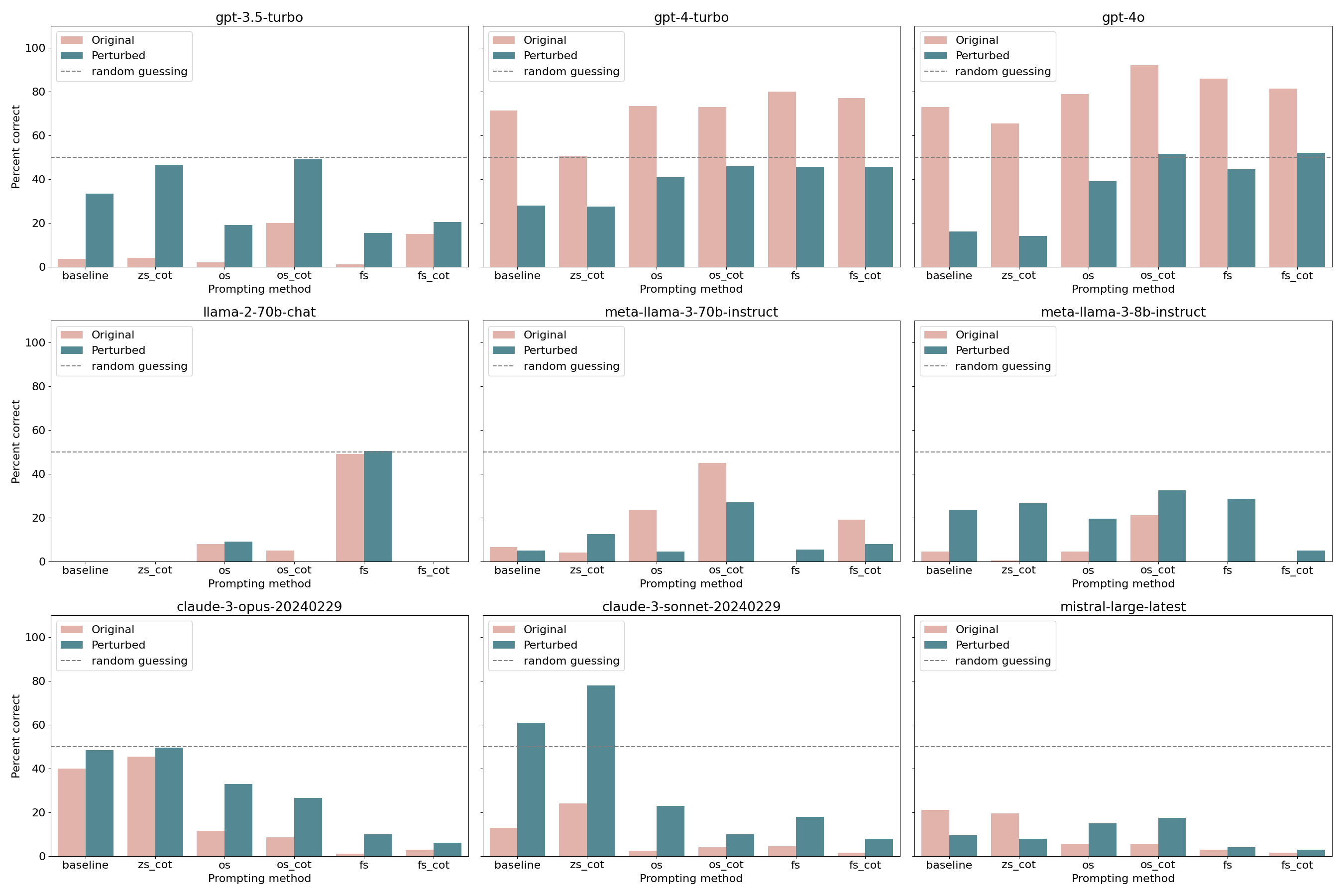}
    \caption{Comparison of the accuracy scores between the original and perturbed problems for Hypothesis \ref{hyp:framing}.}
    \label{fig:h5_full_acc}
\end{figure*}

The full experimental results for Hypothesis \ref{hyp:framing} (sets-original vs. sets-framing) are shown in Figure \ref{fig:h5_full}, \ref{fig:h5_full_acc} and Table \ref{table:h5}.

\begin{longtable}{llrrrrrr}
\caption{Full experimental results for Hypothesis \ref{hyp:framing}(sets-original-random vs sets-original-framing-gold)}
\label{table:h5} \\
\toprule
model & prompting method & $n_{12}$ & $n_{21}$ & $n^*$ & z-stat & p-value & reject \\
\midrule
\endfirsthead

\multicolumn{8}{c}%
{\tablename\ \thetable\ -- \textit{Continued from previous page}} \\
\toprule
model & prompting method & $n_{12}$ & $n_{21}$ & $n^*$ & z-stat & p-value & reject \\
\midrule
\endhead

\midrule \multicolumn{8}{r}{\textit{Continued on next page}} \\
\endfoot

\bottomrule
\endlastfoot

gpt-3.5-turbo & baseline & 1 & 33 & 34 & 5.487955 & 0.000000 & True \\
gpt-3.5-turbo & zs-cot & 2 & 49 & 51 & 6.581316 & 0.000000 & True \\
gpt-3.5-turbo & os & 2 & 18 & 20 & 3.577709 & 0.000836 & True \\
gpt-3.5-turbo & os-cot & 13 & 43 & 56 & 4.008919 & 0.000143 & True \\
gpt-3.5-turbo & fs & 0 & 15 & 15 & 3.872983 & 0.000143 & True \\
gpt-3.5-turbo & fs-cot & 9 & 19 & 28 & 1.889822 & 0.083532 & False \\
gpt-4-turbo & baseline & 47 & 2 & 49 & -6.428571 & 0.000000 & True \\
gpt-4-turbo & zs-cot & 23 & 5 & 28 & -3.401680 & 0.001292 & True \\
gpt-4-turbo & os & 43 & 4 & 47 & -5.688735 & 0.000000 & True \\
gpt-4-turbo & os-cot & 37 & 7 & 44 & -4.522670 & 0.000016 & True \\
gpt-4-turbo & fs & 50 & 4 & 54 & -6.259807 & 0.000000 & True \\
gpt-4-turbo & fs-cot & 46 & 3 & 49 & -6.142857 & 0.000000 & True \\
gpt-4o & baseline & 63 & 2 & 65 & -7.566119 & 0.000000 & True \\
gpt-4o & zs-cot & 57 & 4 & 61 & -6.785955 & 0.000000 & True \\
gpt-4o & os & 50 & 4 & 54 & -6.259807 & 0.000000 & True \\
gpt-4o & os-cot & 53 & 1 & 54 & -7.076304 & 0.000000 & True \\
gpt-4o & fs & 46 & 1 & 47 & -6.563925 & 0.000000 & True \\
gpt-4o & fs-cot & 45 & 5 & 50 & -5.656854 & 0.000000 & True \\
llama-2-70b-chat & baseline & 0 & 0 & 0 & 0.000000 & 1.000000 & False \\
llama-2-70b-chat & zs-cot & 0 & 0 & 0 & 0.000000 & 1.000000 & False \\
llama-2-70b-chat & os & 8 & 8 & 16 & 0.000000 & 1.000000 & False \\
llama-2-70b-chat & os-cot & 5 & 0 & 5 & -2.236068 & 0.086538 & False \\
llama-2-70b-chat & fs & 15 & 21 & 36 & 1.000000 & 0.372495 & False \\
llama-2-70b-chat & fs-cot & 0 & 0 & 0 & 0.000000 & 1.000000 & False \\
meta-llama-3-70b-instruct & baseline & 3 & 4 & 7 & 0.377964 & 1.000000 & False \\
meta-llama-3-70b-instruct & zs-cot & 4 & 10 & 14 & 1.603567 & 0.225501 & False \\
meta-llama-3-70b-instruct & os & 24 & 7 & 31 & -3.053290 & 0.004074 & True \\
meta-llama-3-70b-instruct & os-cot & 31 & 12 & 43 & -2.897473 & 0.006553 & True \\
meta-llama-3-70b-instruct & fs & 20 & 6 & 26 & -2.745626 & 0.010192 & True \\
meta-llama-3-70b-instruct & fs-cot & 15 & 4 & 19 & -2.523573 & 0.028816 & True \\
meta-llama-3-8b-instruct & baseline & 2 & 46 & 48 & 6.350853 & 0.000000 & True \\
meta-llama-3-8b-instruct & zs-cot & 0 & 51 & 51 & 7.141428 & 0.000000 & True \\
meta-llama-3-8b-instruct & os & 3 & 37 & 40 & 5.375872 & 0.000000 & True \\
meta-llama-3-8b-instruct & os-cot & 3 & 51 & 54 & 6.531973 & 0.000000 & True \\
meta-llama-3-8b-instruct & fs & 0 & 53 & 53 & 7.280110 & 0.000000 & True \\
meta-llama-3-8b-instruct & fs-cot & 0 & 10 & 10 & 3.162278 & 0.003637 & True \\
claude-3-opus-20240229 & baseline & 19 & 28 & 47 & 1.312785 & 0.232268 & False \\
claude-3-opus-20240229 & zs-cot & 21 & 28 & 49 & 1.000000 & 0.372495 & False \\
claude-3-opus-20240229 & os & 6 & 29 & 35 & 3.887710 & 0.000228 & True \\
claude-3-opus-20240229 & os-cot & 5 & 23 & 28 & 3.401680 & 0.001292 & True \\
claude-3-opus-20240229 & fs & 1 & 6 & 7 & 1.889822 & 0.168750 & False \\
claude-3-opus-20240229 & fs-cot & 1 & 3 & 4 & 1.000000 & 0.703125 & False \\
claude-3-sonnet-20240229 & baseline & 0 & 49 & 49 & 7.000000 & 0.000000 & True \\
claude-3-sonnet-20240229 & zs-cot & 0 & 54 & 54 & 7.348469 & 0.000000 & True \\
claude-3-sonnet-20240229 & os & 1 & 28 & 29 & 5.013774 & 0.000001 & True \\
claude-3-sonnet-20240229 & os-cot & 2 & 13 & 15 & 2.840188 & 0.011730 & True \\
claude-3-sonnet-20240229 & fs & 2 & 18 & 20 & 3.577709 & 0.000836 & True \\
claude-3-sonnet-20240229 & fs-cot & 1 & 10 & 11 & 2.713602 & 0.018080 & True \\
mistral-large-latest & baseline & 9 & 3 & 12 & -1.732051 & 0.187709 & False \\
mistral-large-latest & zs-cot & 9 & 3 & 12 & -1.732051 & 0.187709 & False \\
mistral-large-latest & os & 2 & 10 & 12 & 2.309401 & 0.056298 & False \\
mistral-large-latest & os-cot & 2 & 13 & 15 & 2.840188 & 0.011730 & True \\
mistral-large-latest & fs & 3 & 4 & 7 & 0.377964 & 1.000000 & False \\
mistral-large-latest & fs-cot & 1 & 3 & 4 & 1.000000 & 0.703125 & False \\
\end{longtable}

\FloatBarrier

\FloatBarrier

\begin{figure*}[h]
    \centering
    \includegraphics[width=\textwidth]{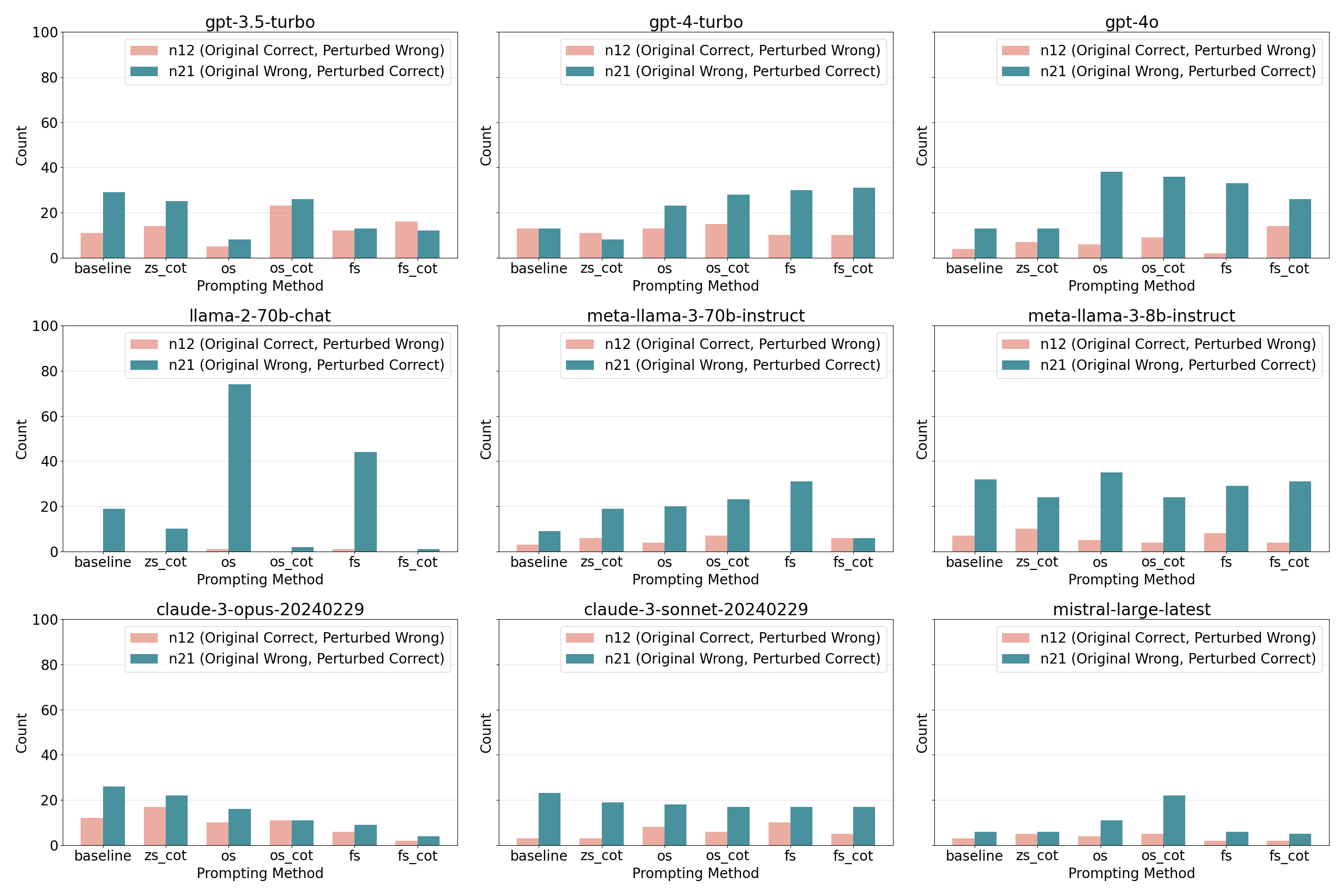}
    \caption{Full experimental results for Hypothesis \ref{hyp:framing}, framing gold vs. random ($n = 200$). The perturbed problems replace reputable news agencies and universities to less trustworthy ones in syllogisms. Both the original and perturbed problems have classic patterns \textit{"All..., Some..., Some..."} already removed to ensure a single confounder for token bias analysis. We run all different prompt methods. To reject the null, we expect $\textcolor{goldenGroupColor}{n12} < \textcolor{controlGroupColor}{n21}.$ We conclude that LLMs might be less likely to be misled by less reputable names. However, such performance shifts should not happen to genuine reasoners, because the names of these entities do not affect the logical essence.}
    \label{fig:h5_full_framing}
\end{figure*}

\begin{figure*}[h]
    \centering
    \includegraphics[width=\textwidth]{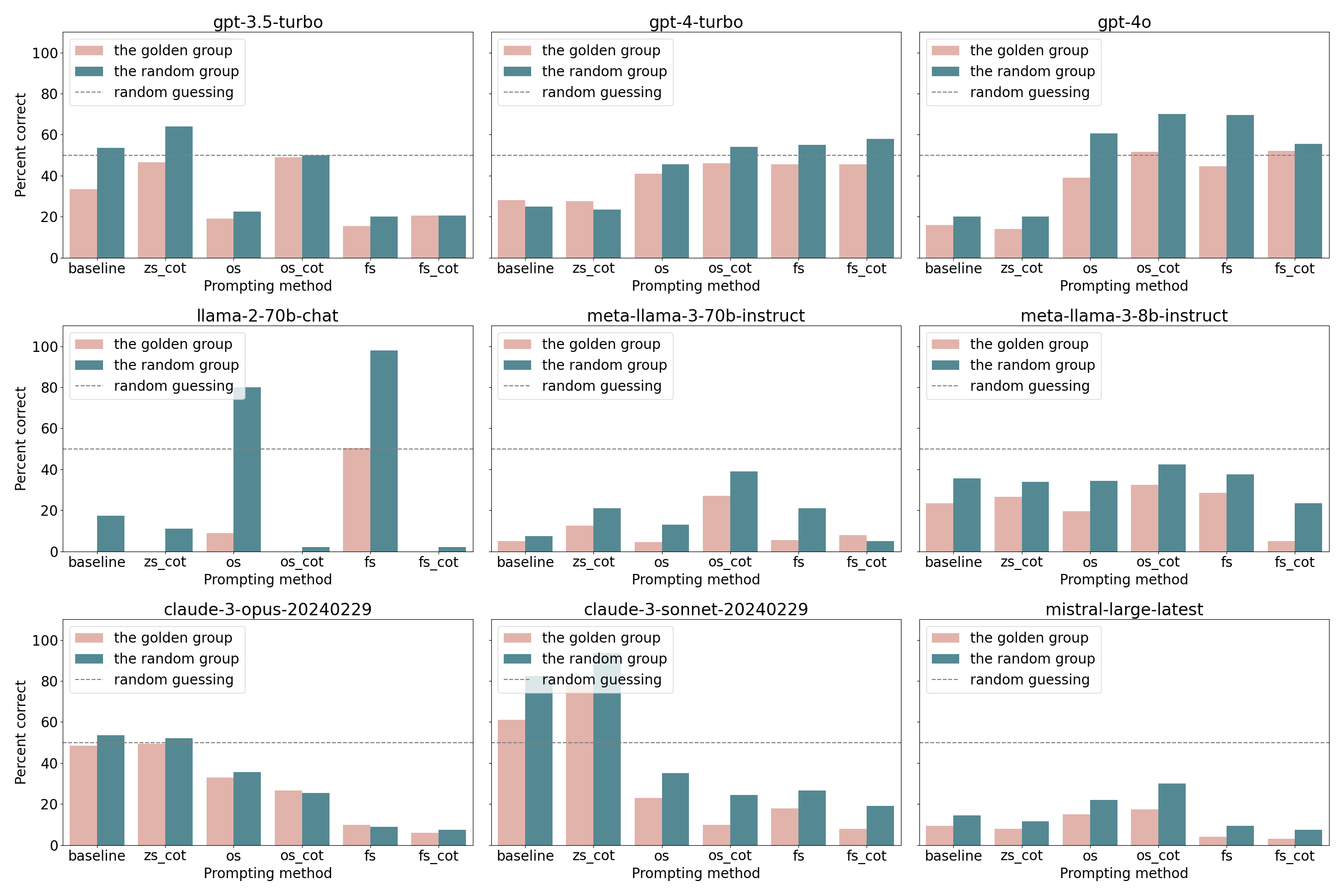}
    \caption{Comparison of the accuracy scores between the original and perturbed problems for Hypothesis \ref{hyp:framing}.}
    \label{fig:h5_full_framing_acc}
\end{figure*}

The full experimental results for Hypothesis \ref{hyp:framing} (framing gold vs. random) are shown in Figure \ref{fig:h5_full_framing},  \ref{fig:h5_full_framing_acc} and Table \ref{table:h5_framing}.

\begin{longtable}{llrrrrrr}
\caption{Full experimental results for Hypothesis \ref{hyp:framing} (sets-framing gold vs. random)}
\label{table:h5_framing} \\
\toprule
model & prompting method & $n_{12}$ & $n_{21}$ & $n^*$ & z-stat & p-value & reject \\
\midrule
\endfirsthead

\multicolumn{8}{c}%
{\tablename\ \thetable\ -- \textit{Continued from previous page}} \\
\toprule
model & prompting method & $n_{12}$ & $n_{21}$ & $n^*$ & z-stat & p-value & reject \\
\midrule
\endhead

\midrule \multicolumn{8}{r}{\textit{Continued on next page}} \\
\endfoot

\bottomrule
\endlastfoot

gpt-3.5-turbo & baseline & 11 & 29 & 40 & 2.846050 & 0.011382 & True \\
gpt-3.5-turbo & zs-cot & 14 & 25 & 39 & 1.761410 & 0.136166 & False \\
gpt-3.5-turbo & os & 5 & 8 & 13 & 0.832050 & 0.713113 & False \\
gpt-3.5-turbo & os-cot & 23 & 26 & 49 & 0.428571 & 0.767760 & False \\
gpt-3.5-turbo & fs & 12 & 13 & 25 & 0.200000 & 0.927346 & False \\
gpt-3.5-turbo & fs-cot & 16 & 12 & 28 & -0.755929 & 0.596799 & False \\
gpt-4-turbo & baseline & 13 & 13 & 26 & 0.000000 & 1.000000 & False \\
gpt-4-turbo & zs-cot & 11 & 8 & 19 & -0.688247 & 0.760233 & False \\
gpt-4-turbo & os & 13 & 23 & 36 & 1.666667 & 0.161292 & False \\
gpt-4-turbo & os-cot & 15 & 28 & 43 & 1.982481 & 0.092843 & False \\
gpt-4-turbo & fs & 10 & 30 & 40 & 3.162278 & 0.004696 & True \\
gpt-4-turbo & fs-cot & 10 & 31 & 41 & 3.279649 & 0.003609 & True \\
gpt-4o & baseline & 4 & 13 & 17 & 2.182821 & 0.092843 & False \\
gpt-4o & zs-cot & 7 & 13 & 20 & 1.341641 & 0.384095 & False \\
gpt-4o & os & 6 & 38 & 44 & 4.824182 & 0.000015 & True \\
gpt-4o & os-cot & 9 & 36 & 45 & 4.024922 & 0.000337 & True \\
gpt-4o & fs & 2 & 33 & 35 & 5.239956 & 0.000002 & True \\
gpt-4o & fs-cot & 14 & 26 & 40 & 1.897367 & 0.104003 & False \\
llama-2-70b-chat & baseline & 0 & 19 & 19 & 4.358899 & 0.000029 & True \\
llama-2-70b-chat & zs-cot & 0 & 10 & 10 & 3.162278 & 0.005551 & True \\
llama-2-70b-chat & os & 1 & 74 & 75 & 8.429314 & 0.000000 & True \\
llama-2-70b-chat & os-cot & 0 & 2 & 2 & 1.414214 & 0.637718 & False \\
llama-2-70b-chat & fs & 1 & 44 & 45 & 6.410062 & 0.000000 & True \\
llama-2-70b-chat & fs-cot & 0 & 1 & 1 & 1.000000 & 1.000000 & False \\
meta-llama-3-70b-instruct & baseline & 3 & 9 & 12 & 1.732051 & 0.231876 & False \\
meta-llama-3-70b-instruct & zs-cot & 6 & 19 & 25 & 2.600000 & 0.022882 & True \\
meta-llama-3-70b-instruct & os & 4 & 20 & 24 & 3.265986 & 0.004696 & True \\
meta-llama-3-70b-instruct & os-cot & 7 & 23 & 30 & 2.921187 & 0.009415 & True \\
meta-llama-3-70b-instruct & fs & 0 & 31 & 31 & 5.567764 & 0.000000 & True \\
meta-llama-3-70b-instruct & fs-cot & 6 & 6 & 12 & 0.000000 & 1.000000 & False \\
meta-llama-3-8b-instruct & baseline & 7 & 32 & 39 & 4.003204 & 0.000337 & True \\
meta-llama-3-8b-instruct & zs-cot & 10 & 24 & 34 & 2.400980 & 0.038026 & True \\
meta-llama-3-8b-instruct & os & 5 & 35 & 40 & 4.743416 & 0.000019 & True \\
meta-llama-3-8b-instruct & os-cot & 4 & 24 & 28 & 3.779645 & 0.000707 & True \\
meta-llama-3-8b-instruct & fs & 8 & 29 & 37 & 3.452379 & 0.002308 & True \\
meta-llama-3-8b-instruct & fs-cot & 4 & 31 & 35 & 4.563833 & 0.000034 & True \\
claude-3-opus-20240229 & baseline & 12 & 26 & 38 & 2.271100 & 0.049984 & True \\
claude-3-opus-20240229 & zs-cot & 17 & 22 & 39 & 0.800641 & 0.586163 & False \\
claude-3-opus-20240229 & os & 10 & 16 & 26 & 1.176697 & 0.358975 & False \\
claude-3-opus-20240229 & os-cot & 11 & 11 & 22 & 0.000000 & 1.000000 & False \\
claude-3-opus-20240229 & fs & 6 & 9 & 15 & 0.774597 & 0.728687 & False \\
claude-3-opus-20240229 & fs-cot & 2 & 4 & 6 & 0.816497 & 0.773438 & False \\
claude-3-sonnet-20240229 & baseline & 3 & 23 & 26 & 3.922323 & 0.000431 & True \\
claude-3-sonnet-20240229 & zs-cot & 3 & 19 & 22 & 3.411211 & 0.003300 & True \\
claude-3-sonnet-20240229 & os & 8 & 18 & 26 & 1.961161 & 0.092843 & False \\
claude-3-sonnet-20240229 & os-cot & 6 & 17 & 23 & 2.293659 & 0.072048 & False \\
claude-3-sonnet-20240229 & fs & 10 & 17 & 27 & 1.347151 & 0.274523 & False \\
claude-3-sonnet-20240229 & fs-cot & 5 & 17 & 22 & 2.558409 & 0.038026 & True \\
mistral-large-latest & baseline & 3 & 6 & 9 & 1.000000 & 0.637718 & False \\
mistral-large-latest & zs-cot & 5 & 6 & 11 & 0.301511 & 1.000000 & False \\
mistral-large-latest & os & 4 & 11 & 15 & 1.807392 & 0.193859 & False \\
mistral-large-latest & os-cot & 5 & 22 & 27 & 3.271652 & 0.003609 & True \\
mistral-large-latest & fs & 2 & 6 & 8 & 1.414214 & 0.410773 & False \\
mistral-large-latest & fs-cot & 2 & 5 & 7 & 1.133893 & 0.596799 & False \\
\end{longtable}

\FloatBarrier

\FloatBarrier

\newpage
\subsection{Hypothesis 6}

\begin{figure*}[h]
    \centering
    \includegraphics[width=\textwidth]{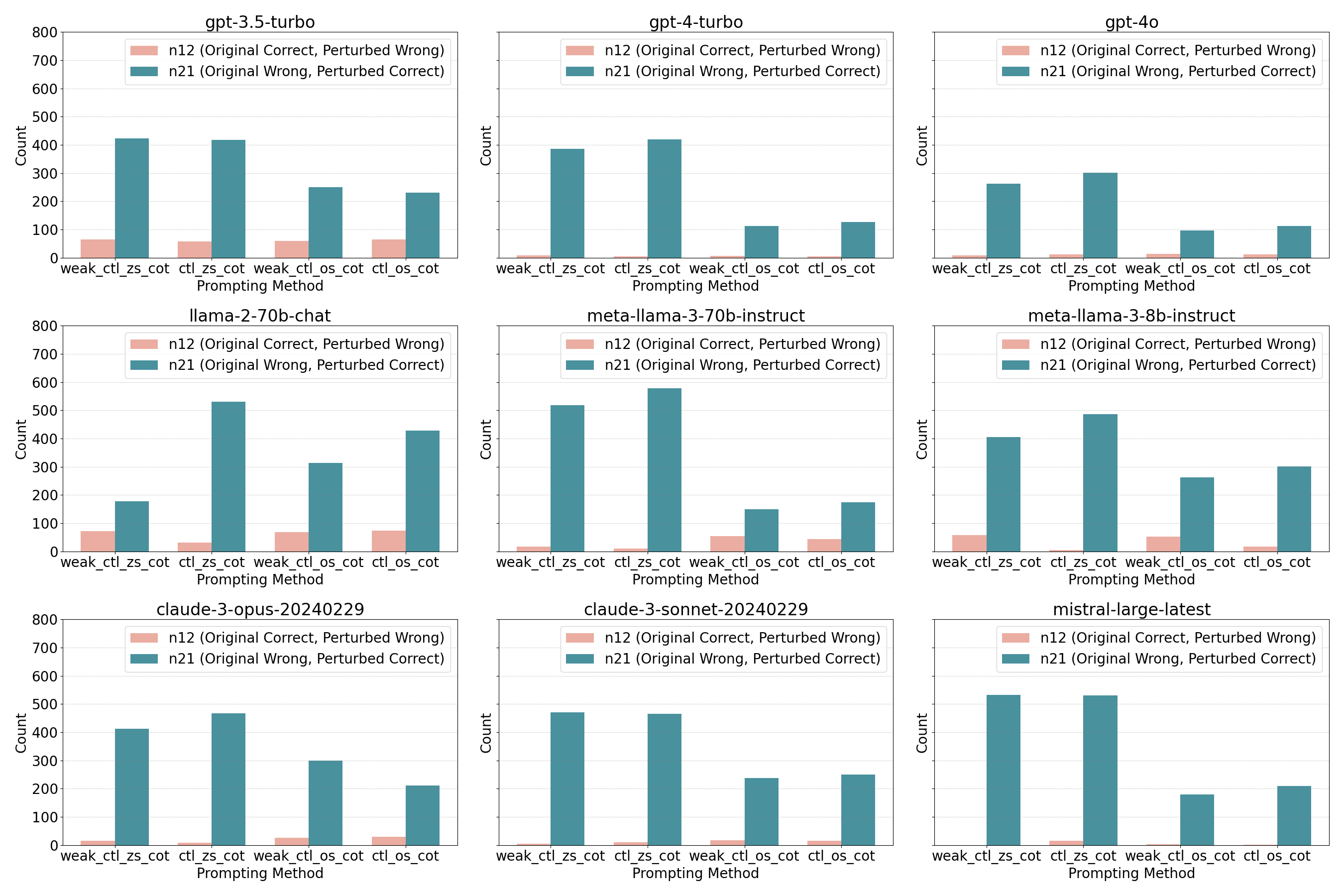}
    \caption{Full experimental results for Hypothesis \ref{hyp:hint} ($n = 800$). The perturbed problems leak hint tokens, either weak or strong hints in problem statements. We run zero-shot and one-shot prompt methods. To reject the null, we expect $\textcolor{goldenGroupColor}{n12} < \textcolor{controlGroupColor}{n21}.$ We conclude that LLMs still heavily rely on hint tokens for solving logical fallacy problems well.}
    \label{fig:h6_full}
\end{figure*}

\begin{figure*}[h]
    \centering
    \includegraphics[width=\textwidth]{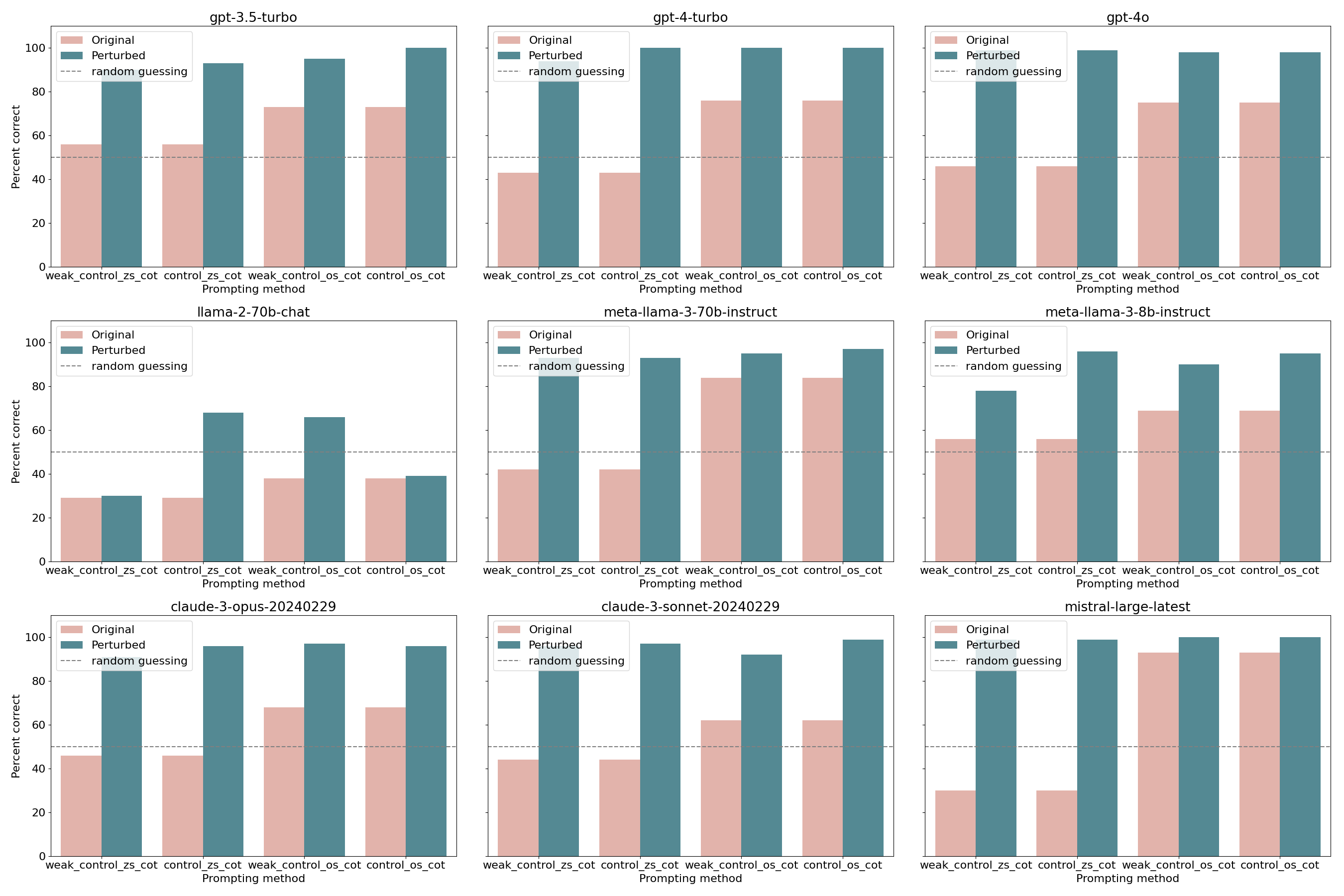}
    \caption{Comparison of the accuracy scores between the original and perturbed problems for Hypothesis \ref{hyp:hint}.}
    \label{fig:h6_full_acc}
\end{figure*}

The full experimental results for Hypothesis \ref{hyp:hint} are shown in Figure \ref{fig:h6_full}, \ref{fig:h6_full_acc} and Table \ref{table:h6}.

\begin{longtable}{llrrrrrr}
\caption{Full experimental results for Hypothesis \ref{hyp:hint} }
\label{table:h6} \\
\toprule
model & prompting method & $n_{12}$ & $n_{21}$ & $n^*$ & z-stat & p-value & reject \\
\midrule
\endfirsthead

\multicolumn{8}{c}%
{\tablename\ \thetable\ -- \textit{Continued from previous page}} \\
\toprule
model & prompting method & $n_{12}$ & $n_{21}$ & $n^*$ & z-stat & p-value & reject \\
\midrule
\endhead

\midrule \multicolumn{8}{r}{\textit{Continued on next page}} \\
\endfoot

\bottomrule
\endlastfoot

gpt-3.5-turbo & weak-control-zs-cot & 64 & 423 & 487 & 16.267843 & 0.000000 & True \\
gpt-3.5-turbo & control-zs-cot & 57 & 417 & 474 & 16.535348 & 0.000000 & True \\
gpt-3.5-turbo & weak-control-os-cot & 60 & 250 & 310 & 10.791275 & 0.000000 & True \\
gpt-3.5-turbo & control-os-cot & 64 & 230 & 294 & 9.681317 & 0.000000 & True \\
gpt-4-turbo & weak-control-zs-cot & 8 & 386 & 394 & 19.043365 & 0.000000 & True \\
gpt-4-turbo & control-zs-cot & 4 & 420 & 424 & 20.202746 & 0.000000 & True \\
gpt-4-turbo & weak-control-os-cot & 6 & 113 & 119 & 9.808674 & 0.000000 & True \\
gpt-4-turbo & control-os-cot & 4 & 126 & 130 & 10.700108 & 0.000000 & True \\
gpt-4o & weak-control-zs-cot & 8 & 262 & 270 & 15.457948 & 0.000000 & True \\
gpt-4o & control-zs-cot & 11 & 301 & 312 & 16.418017 & 0.000000 & True \\
gpt-4o & weak-control-os-cot & 13 & 97 & 110 & 8.009086 & 0.000000 & True \\
gpt-4o & control-os-cot & 12 & 112 & 124 & 8.980265 & 0.000000 & True \\
llama-2-70b-chat & weak-control-zs-cot & 72 & 177 & 249 & 6.654105 & 0.000000 & True \\
llama-2-70b-chat & control-zs-cot & 32 & 531 & 563 & 21.030343 & 0.000000 & True \\
llama-2-70b-chat & weak-control-os-cot & 69 & 313 & 382 & 12.484126 & 0.000000 & True \\
llama-2-70b-chat & control-os-cot & 73 & 429 & 502 & 15.889058 & 0.000000 & True \\
meta-llama-3-70b-instruct & weak-control-zs-cot & 17 & 518 & 535 & 21.660119 & 0.000000 & True \\
meta-llama-3-70b-instruct & control-zs-cot & 10 & 579 & 589 & 23.445237 & 0.000000 & True \\
meta-llama-3-70b-instruct & weak-control-os-cot & 54 & 150 & 204 & 6.721344 & 0.000000 & True \\
meta-llama-3-70b-instruct & control-os-cot & 44 & 174 & 218 & 8.804711 & 0.000000 & True \\
meta-llama-3-8b-instruct & weak-control-zs-cot & 57 & 405 & 462 & 16.190425 & 0.000000 & True \\
meta-llama-3-8b-instruct & control-zs-cot & 4 & 487 & 491 & 21.797485 & 0.000000 & True \\
meta-llama-3-8b-instruct & weak-control-os-cot & 53 & 263 & 316 & 11.813423 & 0.000000 & True \\
meta-llama-3-8b-instruct & control-os-cot & 18 & 301 & 319 & 15.844958 & 0.000000 & True \\
claude-3-opus-20240229 & weak-control-zs-cot & 15 & 412 & 427 & 19.212177 & 0.000000 & True \\
claude-3-opus-20240229 & control-zs-cot & 9 & 467 & 476 & 20.992396 & 0.000000 & True \\
claude-3-opus-20240229 & weak-control-os-cot & 26 & 299 & 325 & 15.143315 & 0.000000 & True \\
claude-3-opus-20240229 & control-os-cot & 30 & 212 & 242 & 11.699403 & 0.000000 & True \\
claude-3-sonnet-20240229 & weak-control-zs-cot & 5 & 470 & 475 & 21.335663 & 0.000000 & True \\
claude-3-sonnet-20240229 & control-zs-cot & 10 & 466 & 476 & 20.900726 & 0.000000 & True \\
claude-3-sonnet-20240229 & weak-control-os-cot & 17 & 238 & 255 & 13.839557 & 0.000000 & True \\
claude-3-sonnet-20240229 & control-os-cot & 16 & 250 & 266 & 14.347461 & 0.000000 & True \\
mistral-large-latest & weak-control-zs-cot & 0 & 533 & 533 & 23.086793 & 0.000000 & True \\
mistral-large-latest & control-zs-cot & 15 & 530 & 545 & 22.060176 & 0.000000 & True \\
mistral-large-latest & weak-control-os-cot & 3 & 179 & 182 & 13.045988 & 0.000000 & True \\
mistral-large-latest & control-os-cot & 1 & 209 & 210 & 14.353364 & 0.000000 & True \\
\end{longtable}

\FloatBarrier

\FloatBarrier

\end{document}